\documentclass[final]{cvpr}

\usepackage{times}
\usepackage{epsfig}
\usepackage{graphicx}
\usepackage{amsmath}
\usepackage{amssymb}

\usepackage{url}            
\usepackage{booktabs}       
\usepackage{amsfonts}       
\usepackage{nicefrac}       
\usepackage{microtype} 
\usepackage{xcolor}
\usepackage{amsmath}
\usepackage{multirow}
\usepackage{subcaption}

\DeclareMathOperator*{\argmin}{argmin}

\usepackage[pagebackref=true,breaklinks=true,colorlinks,bookmarks=false]{hyperref}

\def\cvprPaperID{1300} 
\def\confYear{CVPR 2021}

\begin{document}

\title{Rethinking and Improving the Robustness of Image Style Transfer}

\author{Pei Wang\\
UC, San Diego\\
{\tt\small pew062@ucsd.edu}
\and
Yijun Li\\
Adobe Research\\
{\tt\small yijli@adobe.com}

\and
Nuno Vasconcelos\\
UC, San Diego\\
{\tt\small nuno@ucsd.edu}

}

\maketitle

\begin{abstract}
Extensive research in neural style transfer methods has shown that the correlation between features extracted by a pre-trained VGG network has a remarkable ability to capture the visual style of an image.
Surprisingly, however, this stylization quality is not robust and often degrades significantly when applied to features from more advanced and lightweight networks, such as those in the ResNet family.
By performing extensive experiments with different network architectures, we find that residual connections, which represent the main architectural difference between VGG and ResNet, produce feature maps of small entropy, which are not suitable for style transfer. 
To improve the robustness of the ResNet architecture, we then propose a simple yet effective solution based on a softmax transformation of the feature activations that enhances their entropy.
Experimental results demonstrate that this small magic can greatly improve the quality of stylization results, even for networks with random weights.
This suggests that the architecture used for feature extraction is more important than the use of learned weights for the task of style transfer.
\end{abstract}

\section{Introduction}

\begin{figure}[t]
\centering
\includegraphics[width=0.48\textwidth]{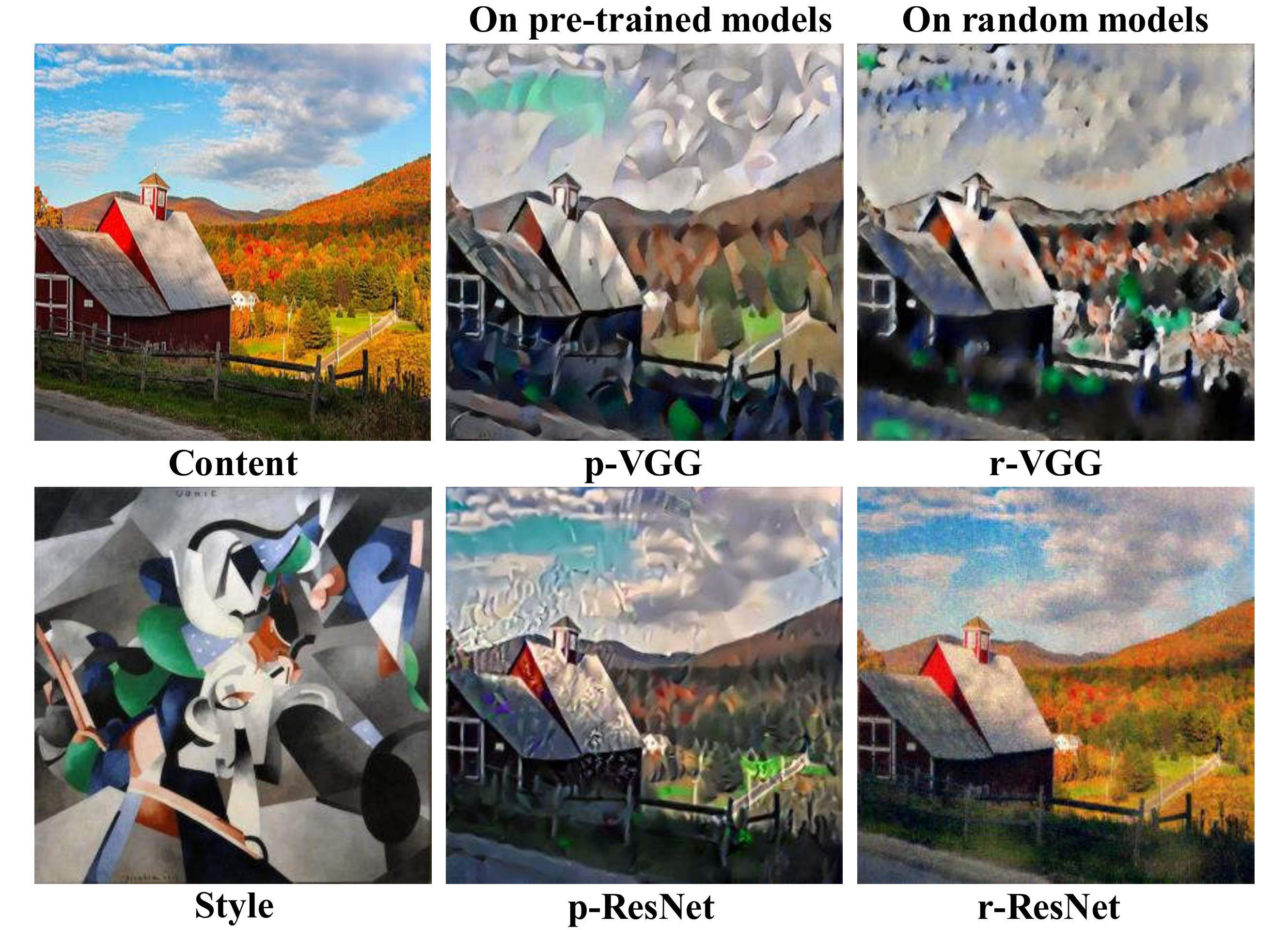}
\caption{Neural style transfer by different architectures, using the methods of \cite{gatys2016image,reiichiro} (`p-', `r-' denotes pre-trained and randomly initialization. Please zoom in the picture for a detailed comparison).}
\label{fig:teaser}
\end{figure}

Image style transfer aims to map a content image into the style of a different reference image.
It has received substantial attention, particularly with the introduction of neural style transfer algorithms based on deep networks. 
%
A consistent observation from this work~\cite{yao2019attention,li2019learning,park2019arbitrary,kolkin2019style,li2018closed,huang2017arbitrary,ghiasi2017exploring,chen2017stylebank,gatys2017controlling} is that the correlation between the activations of a pre-trained VGG~\cite{simonyan2014very} network has remarkable ability to capture the visual style of an image.
%
%
It is, however, puzzling that when the VGG is replaced by architectures of better performance in other tasks, e.g. classification, such as the ResNet~\cite{he2016deep,BMVC2016_87,xie2017aggregated}, InceptionNet~\cite{szegedy2015going,szegedy2017inception,szegedy2016rethinking} or DenseNet~\cite{huang2017densely}, stylization performance degrades significantly. 
This is even more puzzling because, when implemented with the VGG, style transfer is very robust. For example, a VGG model with random weights performs comparably to a pre-trained model~\cite{he2016deep,du2020much}. 
%

Figure~\ref{fig:teaser} shows an example of style transfer using different models. The VGG transfers style (color, texture, strokes) more faithfully than the ResNet, for both pre-trained and random weights. While these observations have spurred discussion in the literature, about why style transfer is much more effective for the VGG~\cite{reiichiro,gwern,engstrom2019a},
%
%
%
there are still no clear answers. In particular, there has not been a comprehensive study of (i) what architectural differences between the VGG and other networks cause this striking performance difference, and (ii) what remedies could make non-VGG networks perform as well as the VGG.
%
%
One explanation is that VGG features are more robust than others. 
To validate this conjecture, \cite{reiichiro,engstrom2019a} trained a ResNet with adversarial examples~\cite{goodfellow2014explaining} to improve feature robustness. 
They found this can significantly improve stylization quality.
%
However, the fact that the VGG with random weights can generate comparable results~\cite{kun2016a,du2020much} suggests that robustness is not a property of the training data, but inherent to the architecture. 

In this work, we investigate this hypothesis by comparing stylizations produced with activations  from different architectures. We seek the architectural properties that explain the differences between these activations, and how these could explain the discrepancy between stylization results. Taking the ResNet as a non-VGG architecture representative, we study the statistics of both activations and the derived Gram matrices, usually used to encode image style.
%
A striking observation is that, when normalized into a probability distribution, the ResNet activations of deeper layers have large peaks and small entropy. This shows that they are dominated by a few feature channels and have nearly deterministic correlation patterns. 
It suggests that the optimization used to synthesize the stylized images is biased into replicating a few dominant patterns of the style image and ignoring the majority. 
This explains why the ResNet is unable to transfer high-level style patterns, such as strokes, that are usually captured in deeper layers of the network.
%
In contrast, VGG activations are approximately uniform for all layers, capturing a much larger diversity of style patterns. 
We then analyze the architectural properties that could lead to very peaky activations, and conclude that they can be, in significant part, explained by the existence of residual or shortcut connections between layers. 
%
The fact that these connections are prevalent in most modern architectures explains why the robustness problem is so widespread. In summary, {\it residual connections are not good for style transfer.\/}

We then investigate whether it is possible to solve the robustness problem without changing the network architecture, and in a manner that is  compatible with the large diversity of stylization losses in the literature. Taking inspiration from knowledge distillation, we propose to smooth the activations used in the computation of these loss functions. This can be implemented by adding a simple softmax transformation to existing losses. We denote the novel version of stylization as {\it Stylization With Activation smoothinG} (SWAG). 
%
Experiments show that SWAG is an important contribution at three levels. First, it improves the performance of several popular stylization algorithms for several popular architectures, including ResNet, Inception, and WideResNet. Second, for these architectures, it improves the performance of random networks to the level of pre-trained ones. Third, for pre-trained networks, non-VGG models with SWAG can even outperform the VGG with standard stylization.



\section{Related Work}

\textbf{Stylization.} For an extensive review of stylization methods please see \cite{jing2019neural}. Starting with \cite{gatys2016image}, it has been shown that a pre-trained image classifier can be used as a feature extractor to drive style transfer~\cite{gatys2016image,johnson2016perceptual,wang2017multimodal}. Style transfer algorithms either implement an iterative optimization~\cite{gatys2016image,du2020much,kun2016a}, or directly learn a feed-forward generator network~\cite{johnson2016perceptual,wang2017stochastic,ulyanov2016texture}. Unlike all these papers, we do not propose a new stylization algorithm. Instead, we note that all previous algorithms use VGG pre-trained models~\cite{yao2019attention,park2019arbitrary,kolkin2019style,li2018closed,huang2017arbitrary,wang2017multimodal,chen2017stylebank,gatys2017controlling,mechrez2018contextual}. In fact, even recent GAN-based proposals~\cite{zhang2017style} use a VGG encoder. We seek to make these algorithms applicable to a broader set of network architectures. To the best of our knowledge, no previous work discusses the architecture robustness of style transfer.

\textbf{Random Networks.} Our work is partly motivated by some theoretical and practical studies on random weights~\cite{huang2004extreme,huang2006extreme,pao1992functional,pao1994learning,wang2017stochastic,wang2017robust,gaier2019weight}. 
For example, Gaier et al.~\cite{gaier2019weight} propose the weight agnostic neural network, by fixing randomly initialized weights and searching optimal network architectures, which is shown to achieve good results in several reinforcement and supervised learning tasks.
\cite{ulyanov2018deep} studies several low-level vision problems and uses random weights to show that the structure of a generator network is sufficient to capture image statistics. \cite{NIPS2018_8160} compares the performance of many saliency algorithms based on random and pre-trained weights.
He et al.~\cite{kun2016a} used a random weight network for texture synthesis and neural style transfer. However, their approach does not  use genuinely random weights, as discussed in Section~\ref{sec:importance}. In contrast, all random models and results presented in this work are based on purely random weights. 

\section{Robust Stylization}


\subsection{Preliminaries} 
Consider a color image $\mathbf{x}_0 \in \mathbb{R}^{W_0 \times H_0 \times 3}$,
where $W_0$ and $H_0$ are the image width and height. A convolutional
neural network (CNN)
maps $\mathbf{x}_0$ into a set of feature maps $\{F^l(\mathbf{x}_0)\}^L_{l=1}$,
where $F^l: \mathbb{R}^{W_0 \times H_0 \times 3} \rightarrow
\mathbb{R}^{W_l \times H_l \times D_l}$ is the mapping from the image
to the tensor of activations of the
$l^{th}$ layer, which has $D_l$ channels of spatial dimensions
$W_l \times H_l$. The activation
tensor $F^l(\mathbf{x}_0)$ can also be reshaped into a matrix
$F^l(\mathbf{x}_0) \in \mathbb{R}^{D_l \times M_l}$, where $M_l = W_l H_l$. 
Image style is frequently assumed to be encoded by a set of Gram matrices $\{G^l\}_{l=1}^L$ where $G^l \in \mathbb{R}^{D_l \times D_l}$ is derived from the activations $F^l$ of layer $l$ by computing the correlation between activation channels, i.e.
\begin{equation}
    [G^l(F^l)]_{ij} = \sum_k F^l_{ik}F^l_{jk}.
\label{equ:gram}
\end{equation}

To simplify the discussion, we focus on the Gram matrix loss~\cite{gatys2016image,li2019learning,ulyanov2016texture,du2020much,kun2016a}.  However, in the experiment section, we show that all results hold for other losses. 
We consider the image stylization framework of \cite{gatys2016image}, where given a content image $\mathbf{x}_0^c$ and a style image $\mathbf{x}_0^s$, an image $\mathbf{x}^*$ that presents the content of $\mathbf{x}_0^c$
under the style of $\mathbf{x}_0^s$ is synthesized by solving
\begin{equation}
     \mathbf{x}^* = \argmin_{x \in \mathbb{R}^{W_0 \times H_0 \times 3}} \alpha \mathcal{L}_{\text{content}}(\mathbf{x}^c_0, \mathbf{x}) + \beta \mathcal{L}_{\text{style}}(\mathbf{x}^s_0, \mathbf{x}).
     \label{equ:stylization_loss}
\end{equation}
with
\begin{align}
  \mathcal{L}_{\text{content}}(\mathbf{x}^c_0, \mathbf{x})
    &= \frac{1}{2}||F^l(\mathbf{x})-F^l(\mathbf{x}_0^c) ||^2_2 \label{equ:content_loss},\\
   \begin{split}
    \mathcal{L}_{\text{style}}(\mathbf{x}^s_0, \mathbf{x}) 
    &= \sum^L_{l=1}\frac{w_l}{4D^2_lM^2_l}||G^l(F^l(\mathbf{x}))-G^l(F^l(\mathbf{x}^s_0))||_2^2, \label{equ:style_loss}
   \end{split}
\end{align}
where $w_l \in \{0, 1\}$ are weighting factors of the contribution of each layer to the total loss. $w_l=1$ represents that $l^{th}$ is used otherwise ignored. $l$ and $w_l$ are pre-specified in (\ref{equ:content_loss}) and (\ref{equ:style_loss}). $\mathbf{x}$ is usually initialized to $\mathbf{x}_0^c$. 

\subsection{Importance of residual connections}
\label{sec:importance}

\begin{figure*}[t]
        \begin{subfigure}[b]{0.199\textwidth}
              \includegraphics[width=\linewidth]{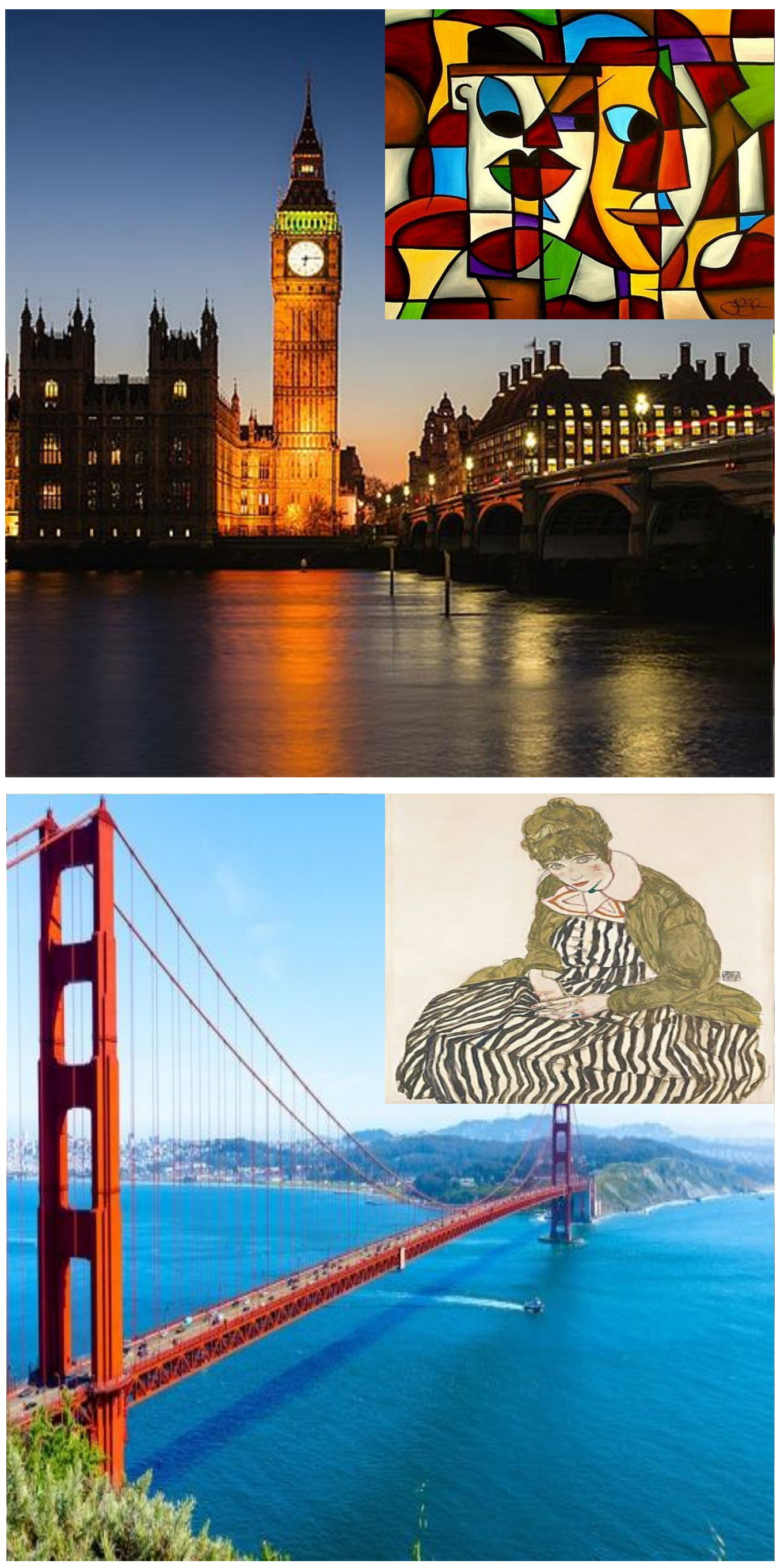}
              \caption{$\text{content}^{\text{style}}$}
                \label{fig:c_s_1}
        \end{subfigure}%
        \begin{subfigure}[b]{0.2\textwidth}
              \includegraphics[width=\linewidth]{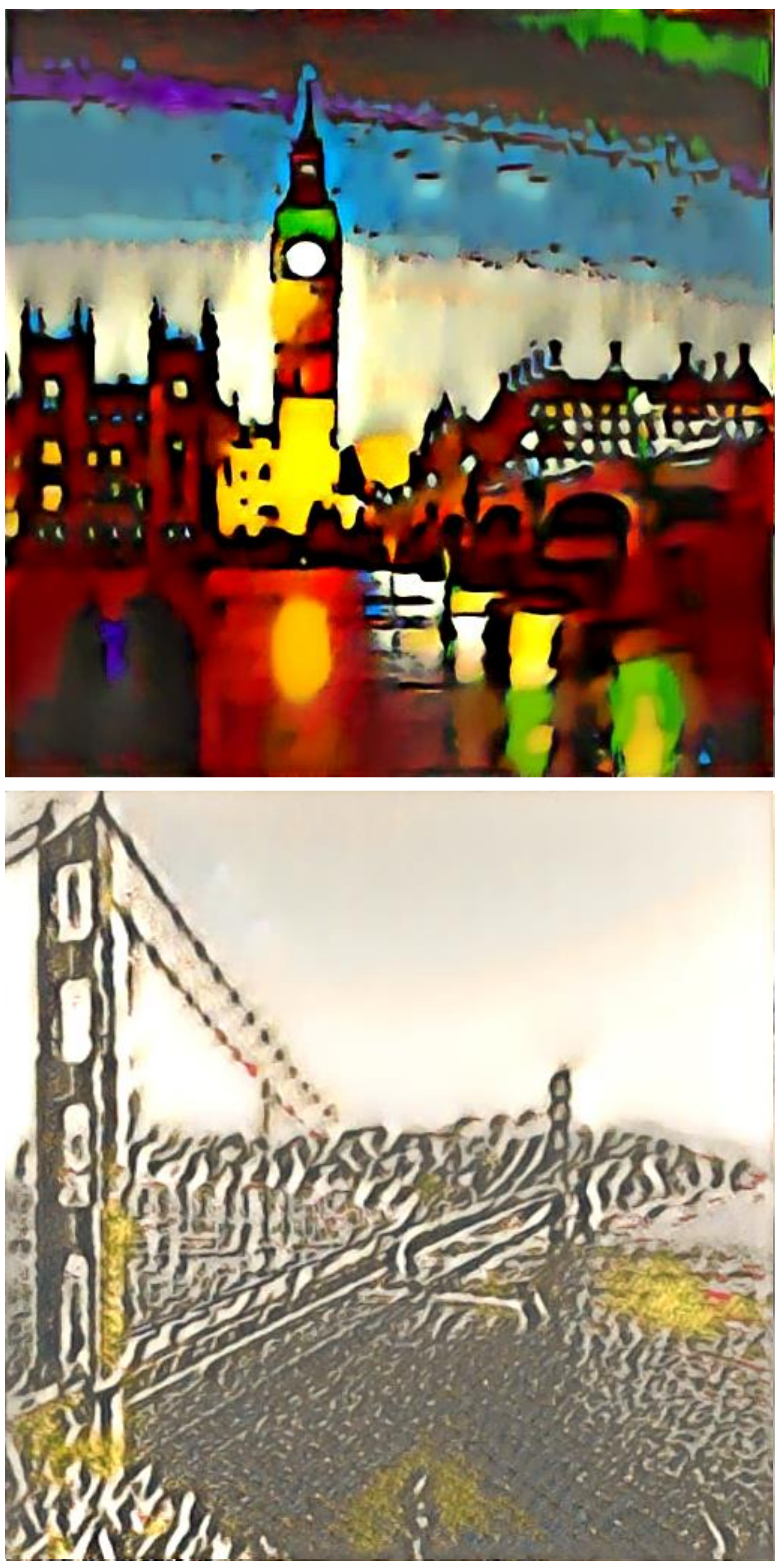}
                \caption{r-VGG}
                \label{fig:r_vgg}
        \end{subfigure}%
        \begin{subfigure}[b]{0.2\textwidth}
              \includegraphics[width=\linewidth]{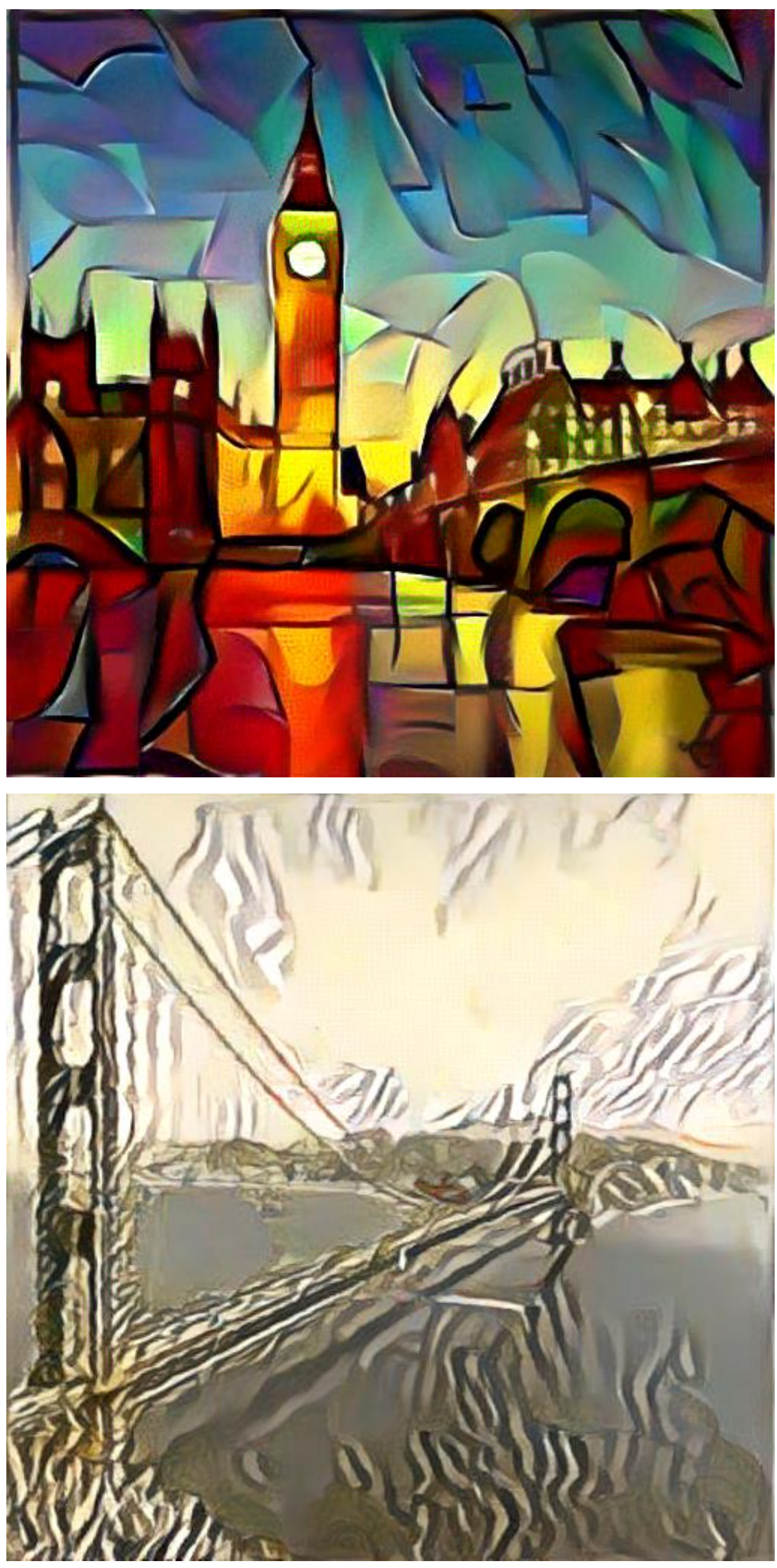}
                \caption{p-VGG}
                \label{fig:pr_vgg}
        \end{subfigure}%
        \begin{subfigure}[b]{0.2\textwidth}
              \includegraphics[width=\linewidth]{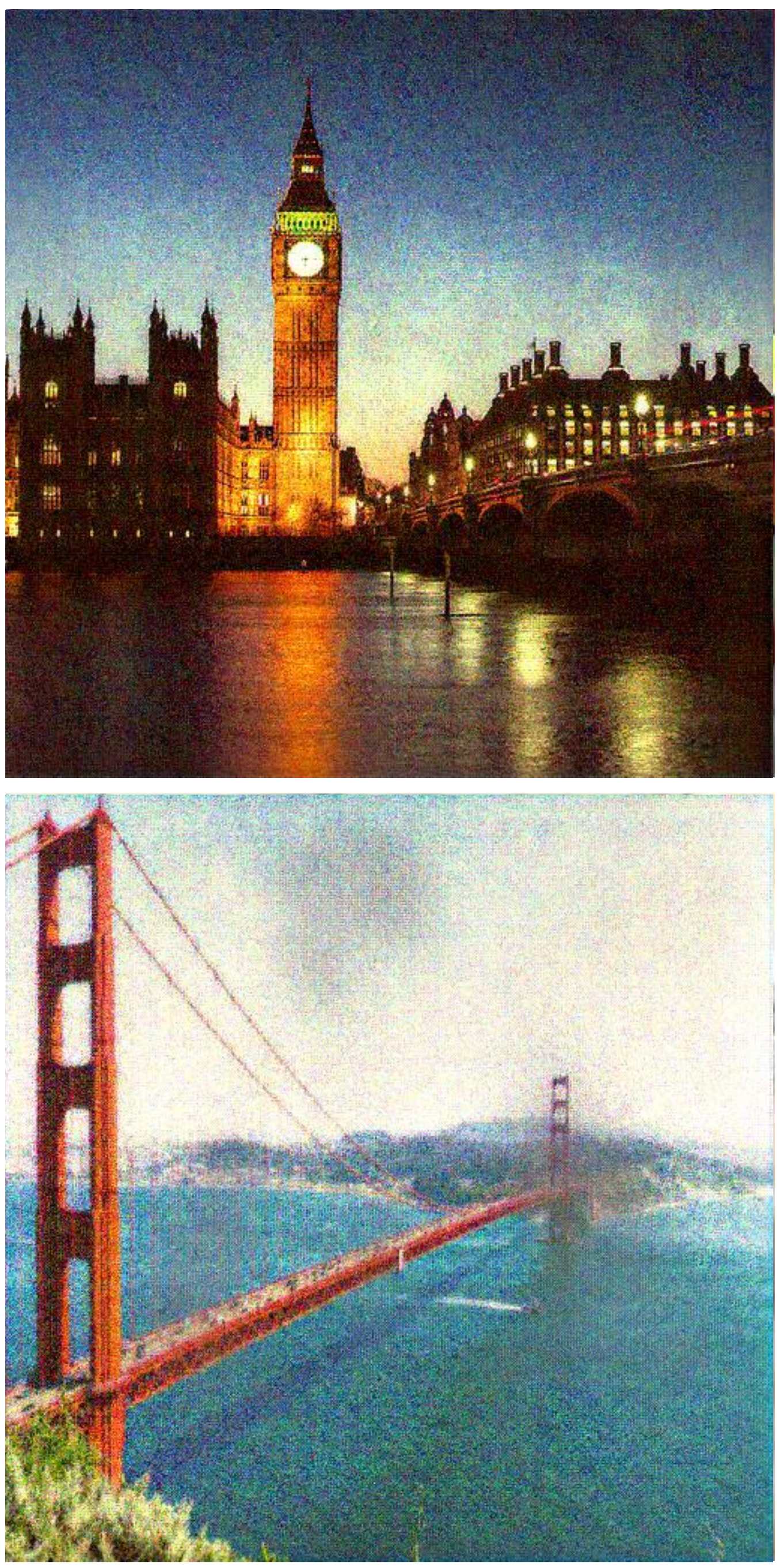}
                \caption{r-ResNet}
                \label{fig:r_resnet}
        \end{subfigure}%
        \begin{subfigure}[b]{0.2\textwidth}
              \includegraphics[width=\linewidth]{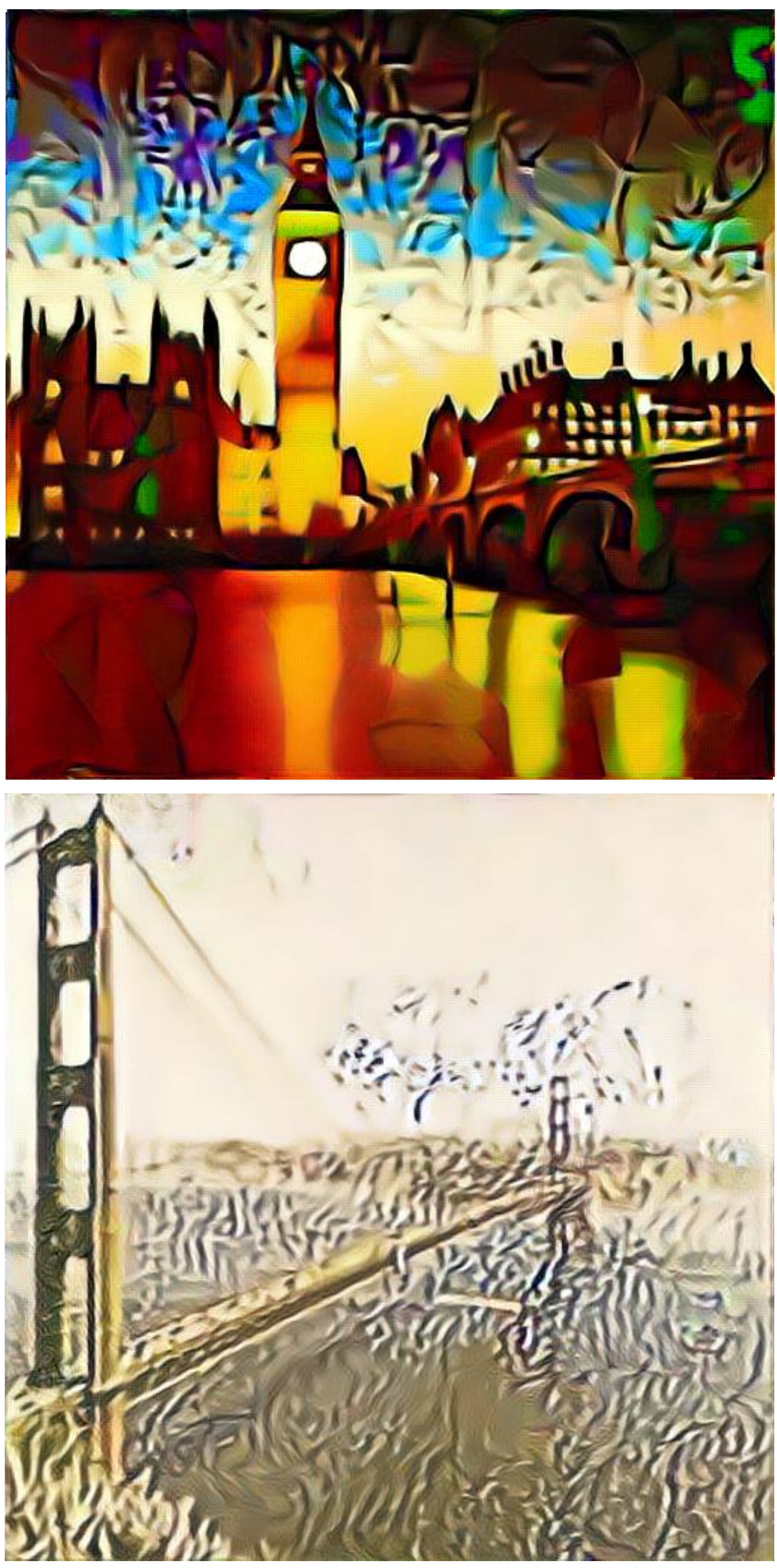}
                \caption{p-ResNet}
                \label{fig:p_resnet}
        \end{subfigure}%
        
        \begin{subfigure}[b]{0.199\textwidth}
              \includegraphics[width=\linewidth]{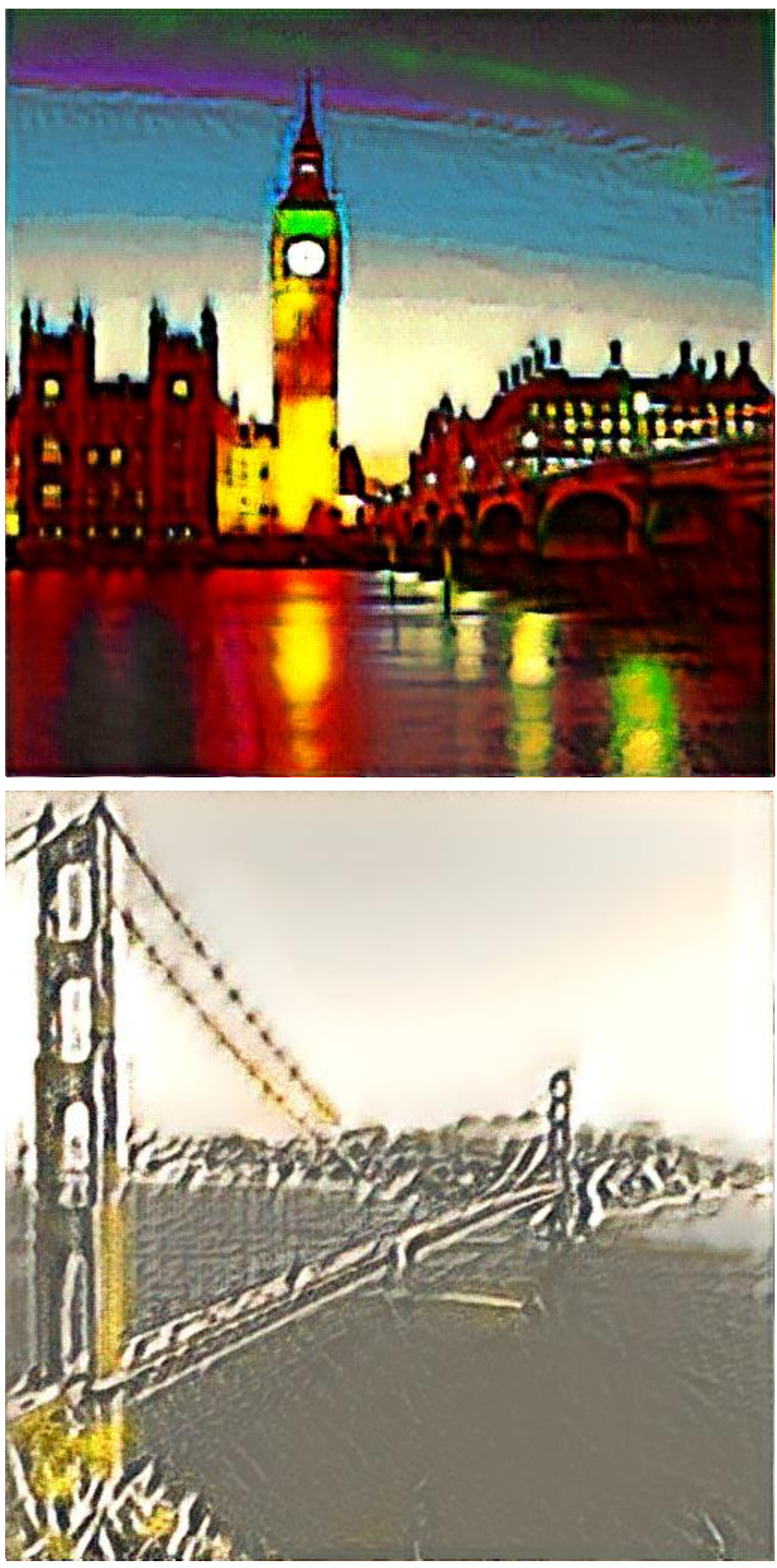}
              \caption{r-NoResNet}
                \label{fig:r_noresnet}
        \end{subfigure}%
        \begin{subfigure}[b]{0.2\textwidth}
              \includegraphics[width=\linewidth]{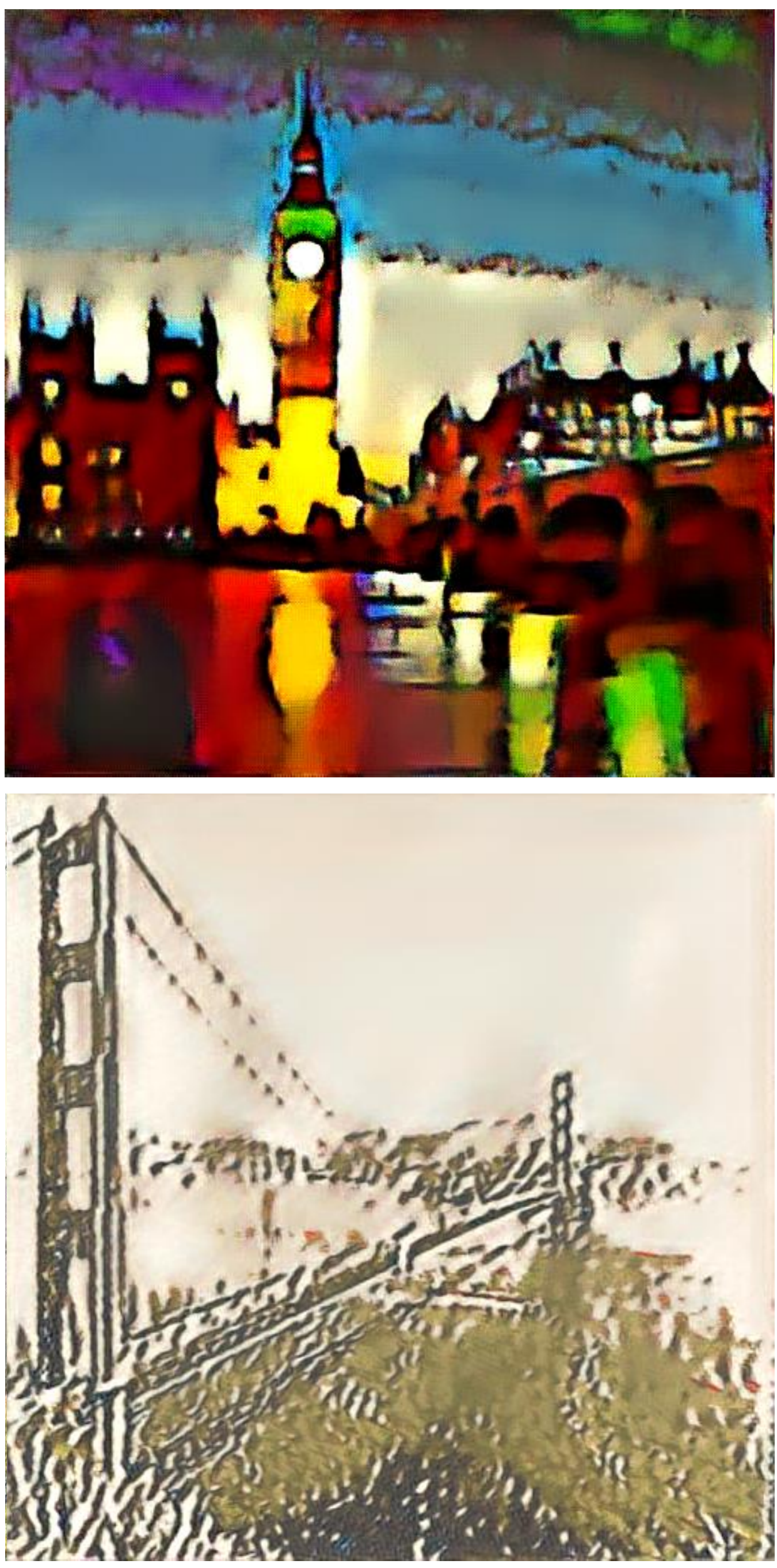}
                \caption{r-pseudo-VGG}
                \label{fig:r_pseudo_vgg}
        \end{subfigure}%
        \begin{subfigure}[b]{0.2\textwidth}
              \includegraphics[width=\linewidth]{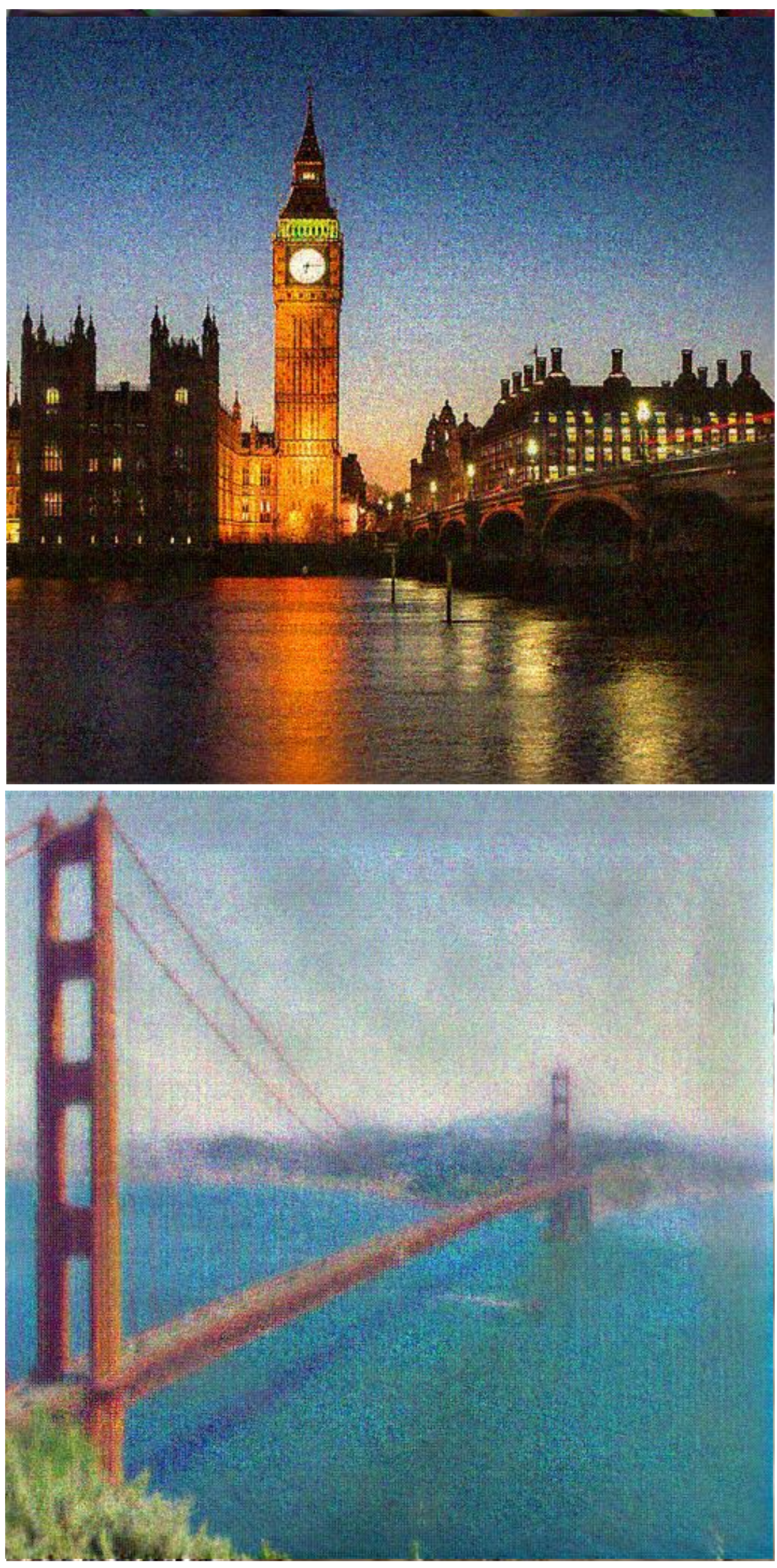}
                \caption{r-pseudo-ResVGG}
                \label{fig:r_pseudo_resvgg}
        \end{subfigure}%
        \begin{subfigure}[b]{0.2\textwidth}
              \includegraphics[width=\linewidth]{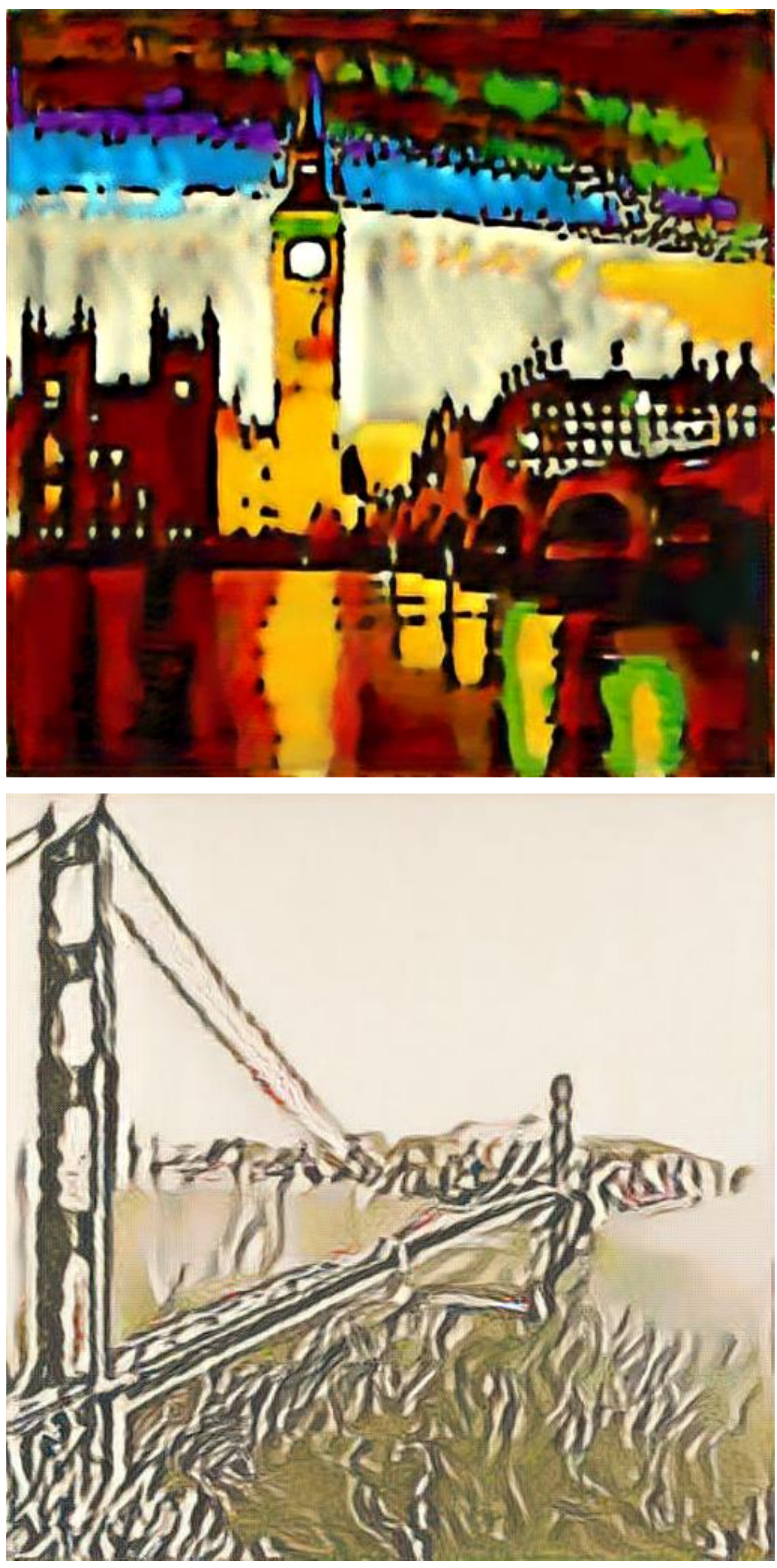}
                \caption{r-ResNet$^*$}
                \label{fig:r_resnet_s}
        \end{subfigure}%
        \begin{subfigure}[b]{0.2\textwidth}
              \includegraphics[width=\linewidth]{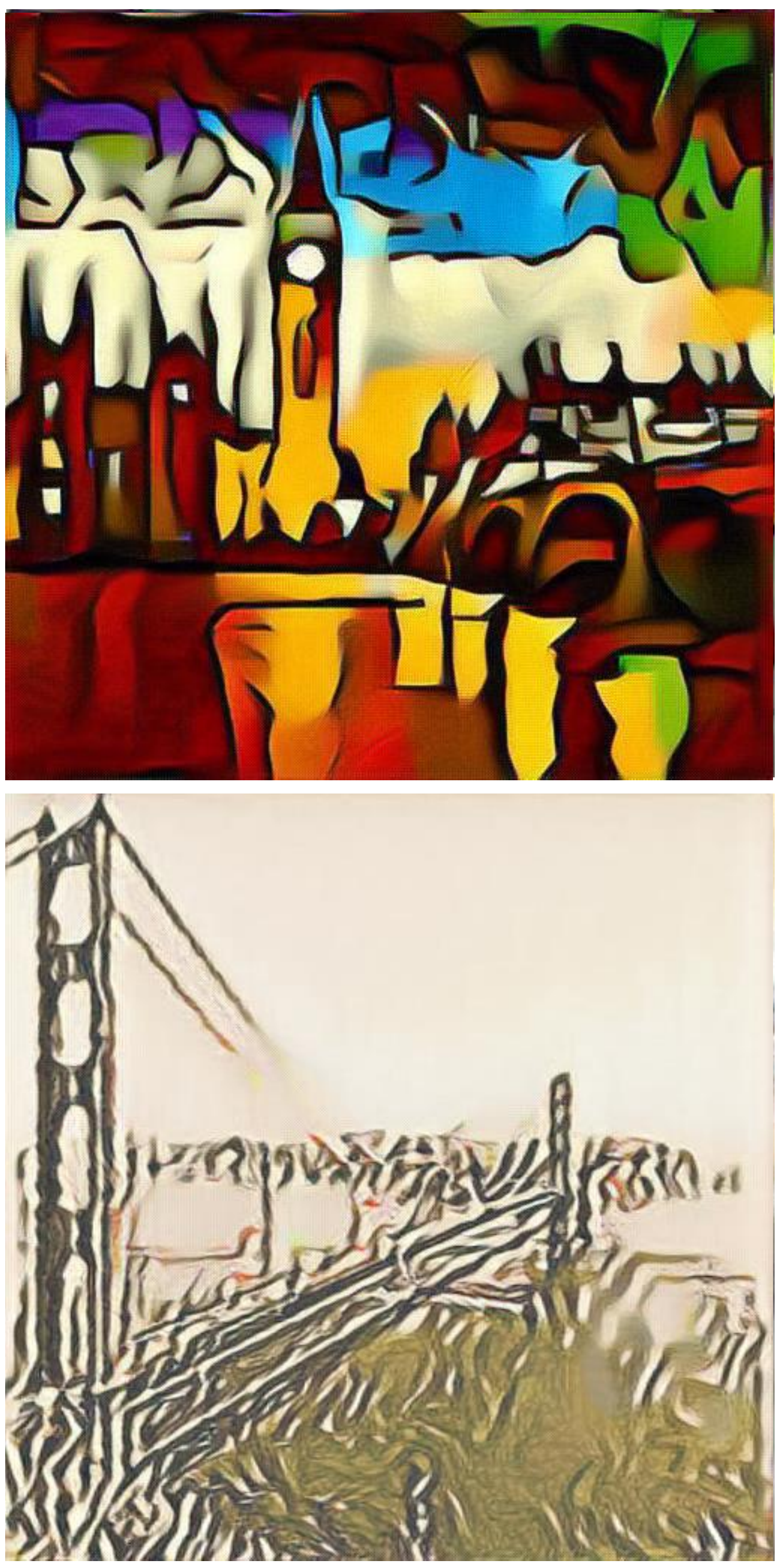}
                \caption{p-ResNet$^*$}
                \label{fig:p_resnet_s}
        \end{subfigure}%
        \caption{Stylization by different architectures. (`p-', `r-' represent pre-trained and randomly initialized, `$^*$' denotes SWAG.)}\label{fig:ablation}
\end{figure*}

We present results for both pre-trained models and networks initialized with random weights. 
Prefixes `r-' and `p-' are used to indicate if a model is initialized randomly or pre-trained on ImageNet, respectively.
We follow the setup of \cite{gatys2016image} for the VGG model, simply replacing their pre-trained models with random networks in some experiments. For ResNet, we follow the setting of \cite{reiichiro}  but, in addition to the outputs of layers conv2$\_$3, conv3$\_$4, conv4$\_$6, conv5$\_$3, we also use that of layer conv1$\_$2 in (\ref{equ:style_loss}). This is for fair comparison to the VGG implementation of~\cite{gatys2016image}, which uses five layers, each selected from one of five layer-groups. VGG19 and ResNet-50 are used as VGG and ResNet models,  because the former is the network of choice for stylization papers and the latter one of the most popular deep learning models.  Random weights follow
default PyTorch settings\footnote{Convolutional layers use
kaiming initialization~\cite{pytorch_vgg,pytorch_resnet}; batchNorm weights (biases) are set to $1$ ($0$); weights of VGG fully
connected layers are drawn from a normal distribution $\mathcal{N}(0, 0.01)$
and biases set to $0$.}. Note that this is unlike
the  random network set-up of \cite{kun2016a}, which samples several sets of random weights
per layer, reconstructs the target image, and chooses the weights yielding the smallest loss\footnote{We believe that this set-up leads to an unfair comparison to \cite{gatys2016image}, since it 
leverages the target image to choose the best random weights and extremely time-consuming gradient computations to add weight factors in (\ref{equ:style_loss}).}~\cite{kun2016a,du2020much}.

Figures \ref{fig:r_vgg} to \ref{fig:p_resnet} present two examples of stylization by the r-VGG, p-VGG, r-ResNet and p-ResNet networks, showing that performance varies drastically with the network architecture. Compared to p-VGG, the p-ResNet transfers much lower-level color patterns and produces much noisier stylized images. This discrepancy is even more obvious for random models, with the r-ResNet simply failing to stylize the content image.
To investigate the reasons behind the very different performance of two architectures, we performed an ablation study over many network components, including the use of residual connections, convolution kernels varying among
size $1 \times 1$, $3 \times 3$ to $7 \times 7$, variable network depth, batch normalization, number of channels per layer, a fixed stride of $2$ vs. maxpooling, etc. While detailed results are presented in the supplementary, the main conclusion was that the poor performance of the ResNet is mainly explained by its residual connections. This is
interesting, since residual connections are usually seen as the main asset of  this architecture for tasks like classification.

Figure~\ref{fig:ablation} provides evidence for this claim,
comparing the stylized outcomes by several architectures after deletion or addition of residual connections. Starting
from the ResNet-50, we built a `NoResNet' by removing all residual connections. As can be seen in Figure~\ref{fig:r_noresnet}, this drastically improves style transfer performance. r-NoResNet has much closer performance to r-VGG than to r-ResNet. We next considered the benefits of several other modifications that made the NoResNet more similar to
the VGG: 1) replaced its $7\times 7$ conv kernel with the $3\times 3$ conv kernel of the VGG; 2) replaced the bottleneck module with the basicblock module of the ResNet-34
(see Figure~\ref{fig:bottleneck}) without the residual connection; 3) inserted a maxpooling layer between each stage to decrease the size of feature maps, as done in the VGG.  The resulting architecture is denoted as 
'pseudo-VGG'. As shown in Figure \ref{fig:r_pseudo_vgg}, these modifications made the stylization performance even closer to that of the r-VGG. However, by comparing
the r-ResNet, r-NoResNet, and r-pseudo-VGG stylizations, it is clear that the bulk of the gains are due to the deletion of residual connections, i.e., from ResNet to NoResNet. To further confirm this, we re-introduced the residual connections in the pseudo-VGG network, to create a 'pseudo-ResVGG'. Figure~\ref{fig:r_pseudo_resvgg} shows that this destroyed all the gains of the pseudo-VGG, again producing undiscernable styles. In fact, this network produced the worst results of Figure \ref{fig:ablation}, showing that there is no benefit
to the pseudo-ResVGG over the ResNet. In summary, the weak ResNet performance is due to its residual connections.

\subsection{Why do residual connections degrade performance?}


\begin{figure*}
\centering
\begin{minipage}{.8\textwidth}
          \begin{subfigure}[b]{0.25\textwidth}
              \includegraphics[width=\linewidth]{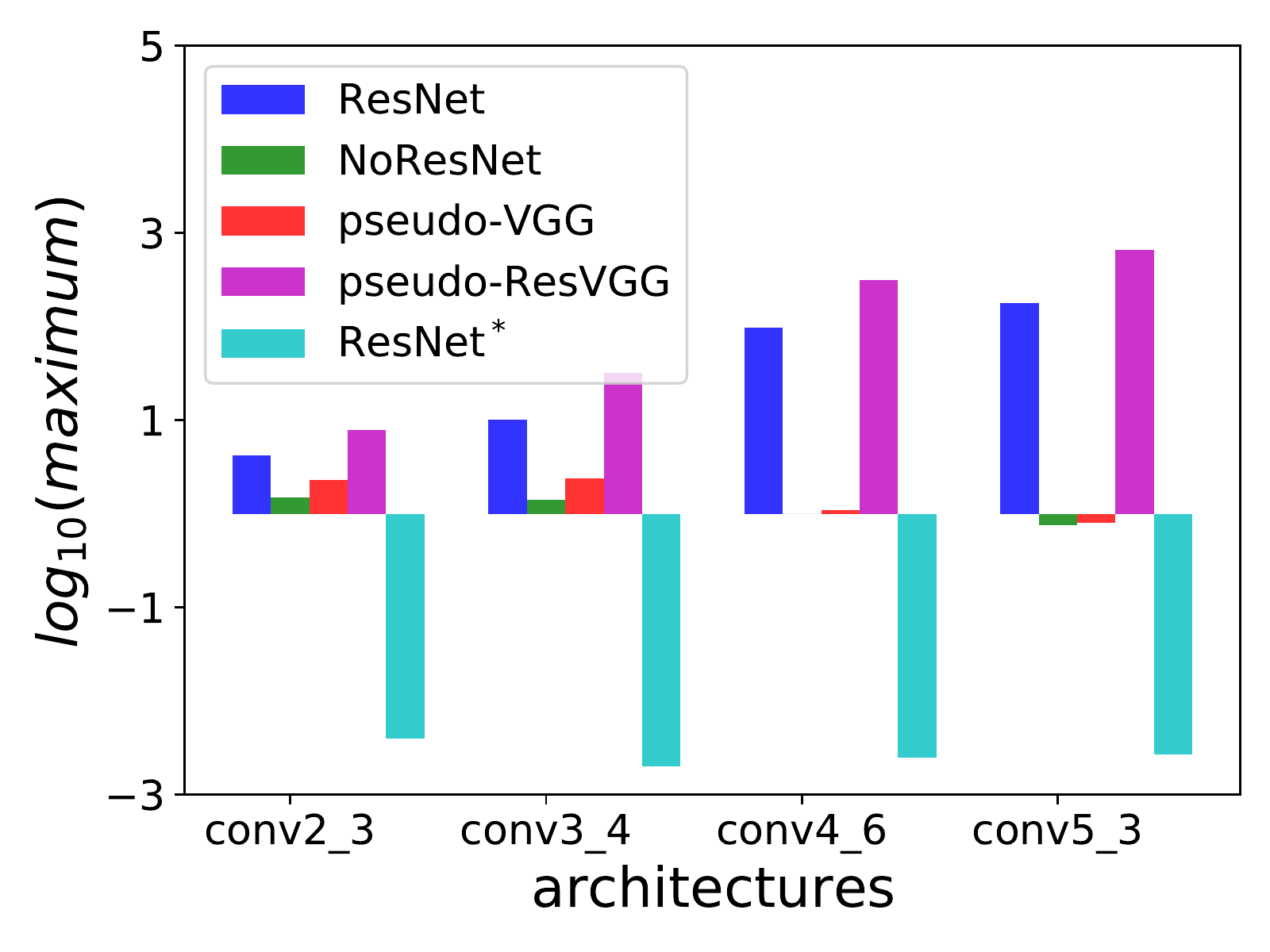}
              \caption{Activation maxima.}
                \label{fig:comp_max}
        \end{subfigure}%
        \begin{subfigure}[b]{0.245\textwidth}
              \includegraphics[width=\linewidth]{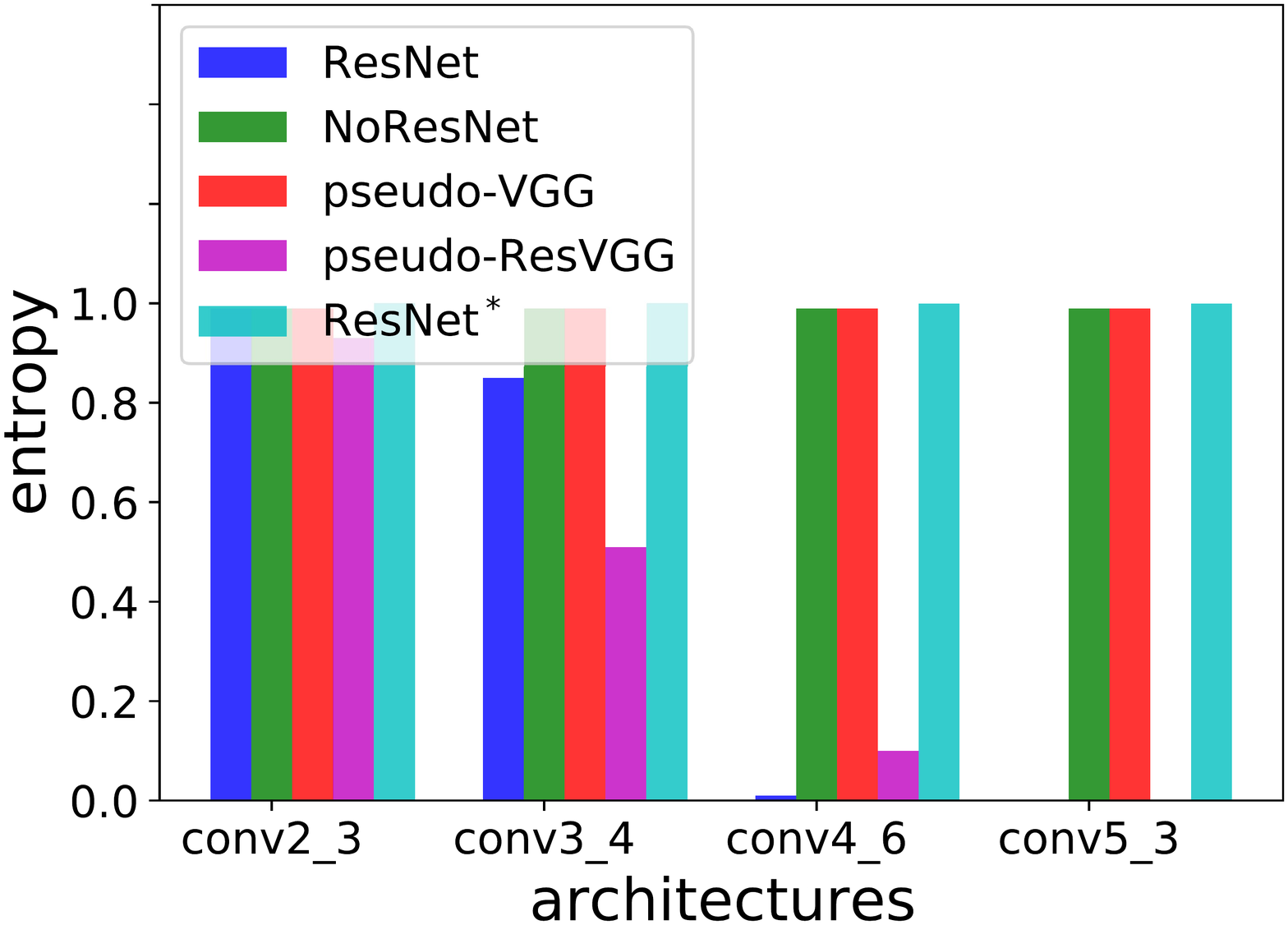}
                \caption{Activation entropy.}
                \label{fig:comp_entropy}
        \end{subfigure}%
        \begin{subfigure}[b]{0.25\textwidth}
              \includegraphics[width=\linewidth]{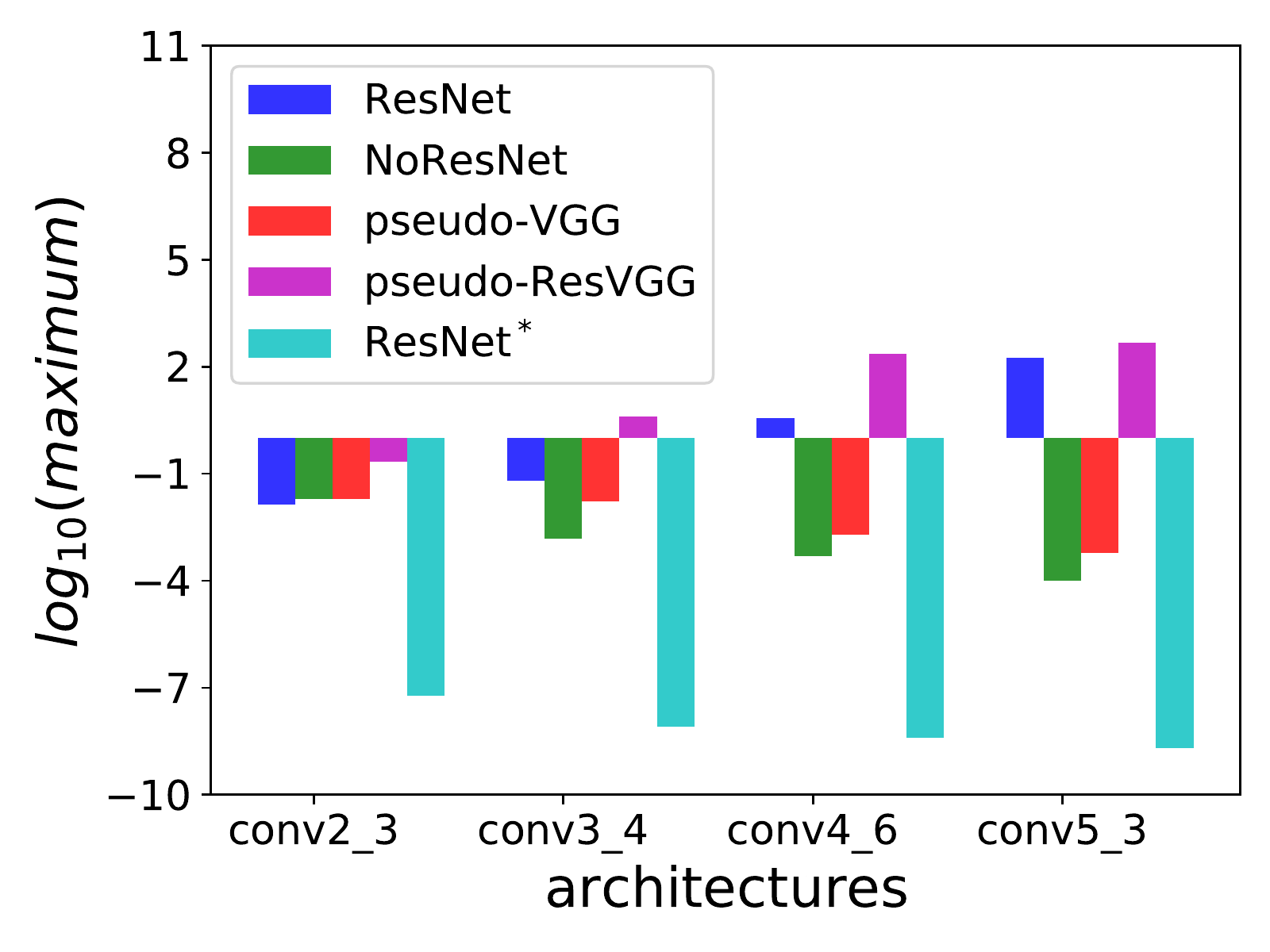}
                \caption{Gram maxima.}
                \label{fig:comp_max_gram}
        \end{subfigure}%
        \begin{subfigure}[b]{0.245\textwidth}
              \includegraphics[width=\linewidth]{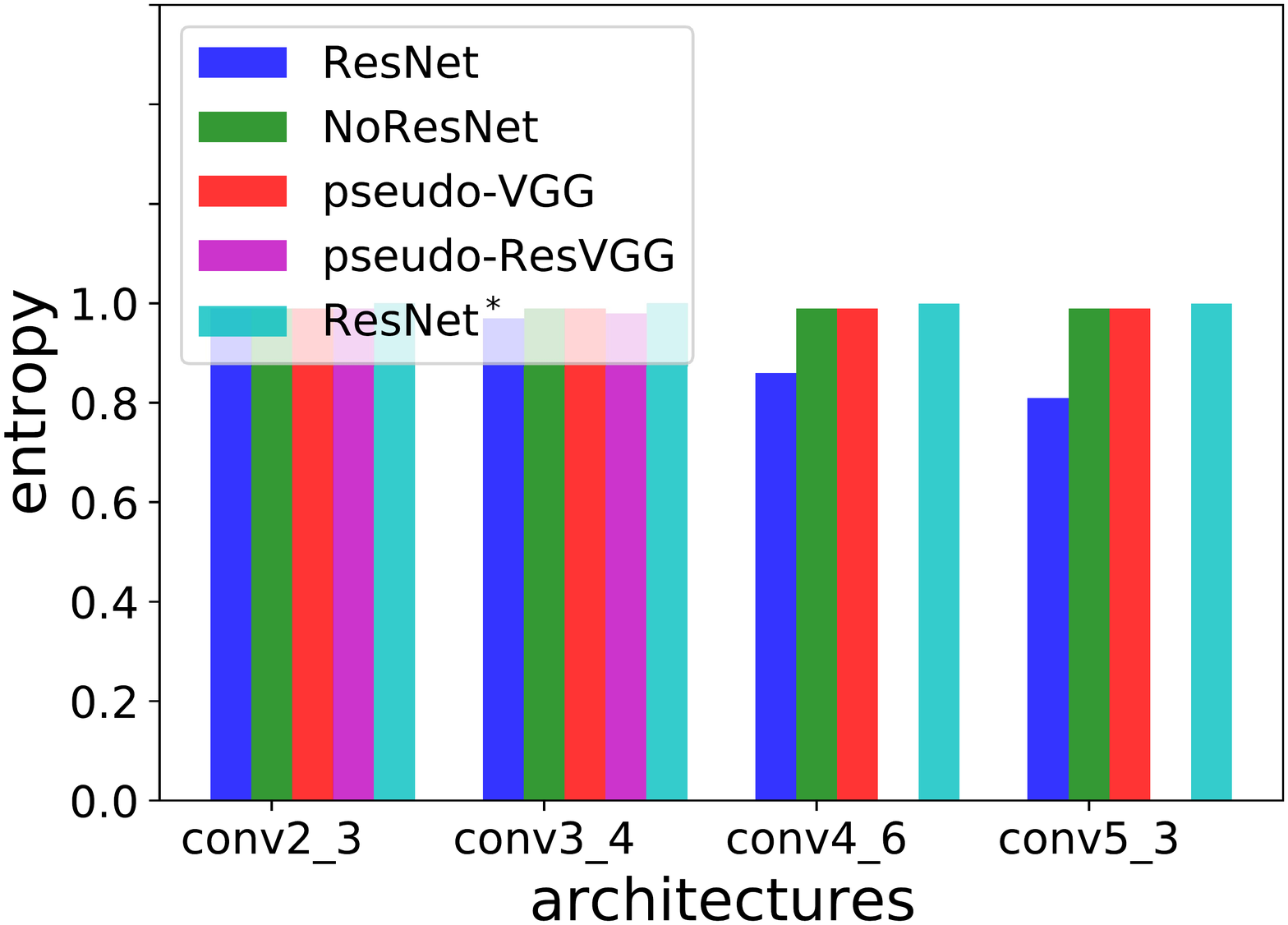}
                \caption{Gram entropy.}
                \label{fig:comp_entropy_gram}
        \end{subfigure}%
        \caption{Activation statistics of different random architectures.}\label{fig:feature}
\end{minipage}%
\begin{minipage}{.2\textwidth}
  \centering
  \includegraphics[width=1\textwidth]{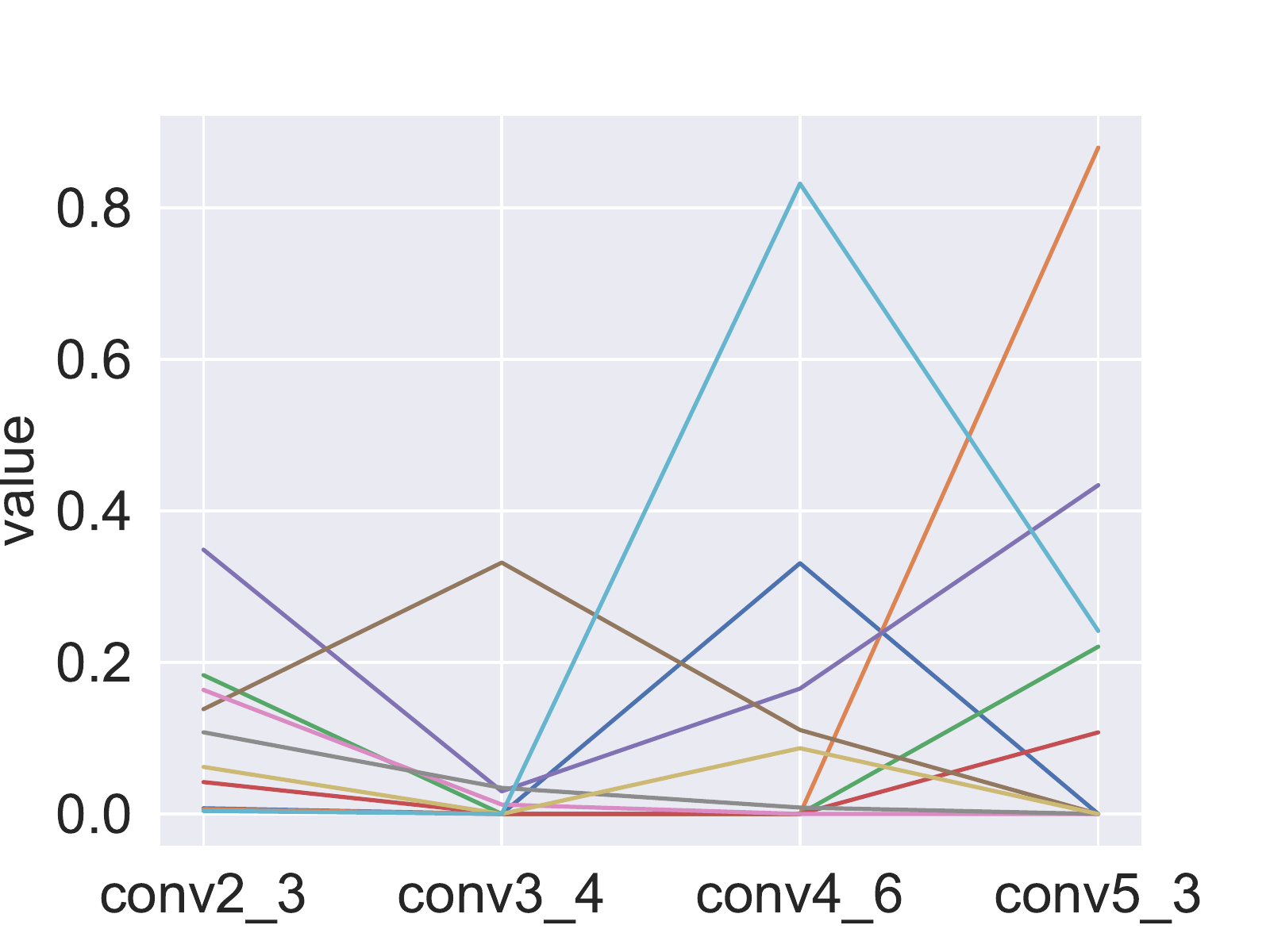}
 \caption{Activation tracks across layers.}
 \label{fig:value_evalution}
\end{minipage}
\end{figure*}


We next try to understand why residual connections are so nefarious for stylization.
Since the optimization of (\ref{equ:style_loss}) is, in
this case, solely based on the Gram matrices $G^l$ of the network responses
to the original and synthesized style, 
we start by visualizing the statistics of the network activations and their Gram matrices. 
Figure \ref{fig:feature} presents the average of maximum values $\max_{i,k} F_{i,k}^l$ and $\max_{i,j} G_{i,j}^l$, as well as normalized entropies~\cite{gray2011entropy}\footnote{We omit the dependency on $\mathbf{x}$ on these equations for simplicity.} 
\begin{eqnarray}
    &H(F^l_{i,k}) = -\frac{1}{\log (D_l M_l)^2}\sum_{i,k} p(F^l_{i,k}) \log p(F^l_{i,k})\\ 
    &p(F^l_{i,k}) = \frac{e^{F^l_{i,k}}}{\sum_{m,n}e^{F^l_{m,n}}}
    \label{eq:fentropy} \\
    &H(G^l_{i,j}) = -\frac{1}{\log D_l^2}\sum_{i,j} p(G^l_{i,j}) \log p(G^l_{i,j}),\\
    &p(G^l_{i,j}) = \frac{e^{G^l_{i,j}}}{\sum_{m,n}e^{G^l_{m,n}}}
\end{eqnarray}
of the activations and Gram matrices, respectively, of the last layer of each network stage of the random models on $10$ style images, where $i,j,k$ are the spatial coordinates defined in (\ref{equ:gram}).
The figure shows that activations and Gram values have similar behavior. In both cases, the maximum value increases and the entropy decreases gradually with layer depth for the architectures with residual connections (ResNet and pseudo-ResVGG). This is unlike the networks without shortcuts (NoResNet and pseudo-VGG), where activations tend to decrease and entropies remain fairly constant and much higher. 
In some cases, e.g. pseudo-ResVGG, the introduction of residual connections translates into large maximum and near-zero entropies for the deeper layers, i.e. activations that are dominated by a single feature channel and deterministic correlation patterns.

The observation of small entropies is consistent with at least two explanations for poor stylization performance. Both of these follow from the fact that the only variables in the optimization of (\ref{equ:style_loss}) are the activations $F^l(\mathbf{x})$, which are encouraged to have Gram matrix as similar as possible to those of the style image $F^l(\mathbf{x}^s_0)$. A first explanation is derived from the well known outlier sensitivity of the $L_2$ distance~\cite{huber2004robust}. Due to this outlier sensitivity, when the Gram matrices derived from
$\mathbf{x}^s_0$ are ``peaky'' (low entropy), the optimization will focus mostly on creating equal peaks on the matrices produced by $\mathbf{x}^*$, while paying less attention to the remaining entries of the Gram matrix.
From (\ref{equ:gram}), the Gram matrix value of location $i,j$ is a measure of similarity of the vectors of activations $F^l_{i,k}$ and $F^l_{j,k}$ across the channel depth dimension $k$. Hence, the Gram matrix peaks identify pairs of locations of strong activations that are highly correlated across the channel dimension. By giving
disproportionate emphasis to these location pairs, the optimization overfits on a few style patterns, ignoring most of the remaining. This explains why small entropies degrade stylization. Note that the appearance of Gram peaks in the deeper layers is consistent with the stylization results of Figure~\ref{fig:ablation}. While the r-ResNet and r-pseudo-ResVGG can transfer the localized color patterns captured by the earlier layers, they fail to capture the long-range correlations that are essential for texture and style perception and only accessible in the later layers.

A second explanation derives from interpreting stylization as a knowledge distillation problem~\cite{hinton2015distilling}.
For classification, neural networks 
are typically trained to minimize a cross-entropy loss between the posterior distribution $\mathbf{q}$ and a target distribution $\mathbf{p}$,
which is typically a one-hot code. Knowledge distillation aims to, instead, minimize the distance between the distribution of a student network $\mathbf{q}$ and that of a soft teacher network $\mathbf{p}$. \cite{hinton2015distilling} has shown that using the soft probability output of a pre-trained larger network as target $\mathbf{p}$ can improve training speed and converge to a better model than using the hard one-hot target. This is because a teacher distribution of high entropy produces gradients of much less variance during training.
The same rationale can be applied to stylization, which can be seen as a form of knowledge distillation, where
the optimization of (\ref{equ:stylization_loss}) seeks the $\mathbf{x}^*$ that minimizes the distance
between the `student' $(F^l(\mathbf{x})$, $G^l(F^l(\mathbf{x})))$ and 
`teacher' $(F^l(\mathbf{x}^c_0), G^l(F^l(\mathbf{x}^s_0)))$ pairs of activations and Gram matrices. Under the distillation view, pairs of higher entropy should make learning easier.

\subsection{Why are residual network activations and Gram matrices peaky?}

\begin{figure}[t]
 \centering
  \includegraphics[width=0.45\textwidth]{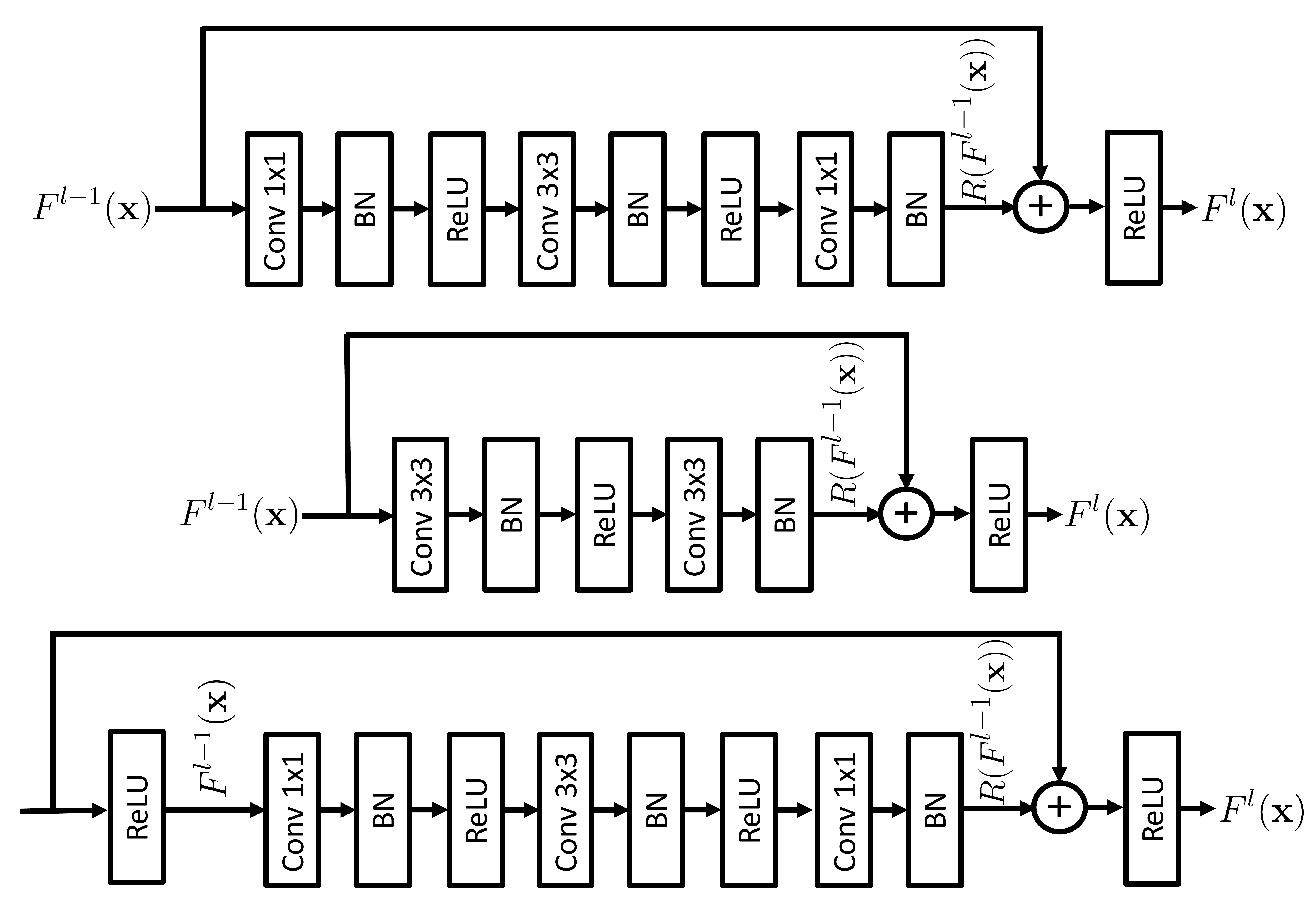}
 \caption{Top: bottleneck; Middle: Basicblock; Bottom: a variation of bottleneck.}
 \label{fig:bottleneck}
 \end{figure}

We next investigate why residual networks produce peaky activations and Gram matrices. We start by recalling the details of the bottleneck ResNet block, shown in Figure \ref{fig:bottleneck}. At layer $l$, this module computes a residual $R(F^{l-1}(\mathbf{x}))$, which is added to the output $F^{l-1}(\mathbf{x})$ of layer $l-1$ to create the output of layer $l$
\begin{equation}
    F^l(\mathbf{x}) = R(F^{l-1}(\mathbf{x})) + F^{l-1}(\mathbf{x}),
    \label{equ:identity1}
\end{equation}
where $R(\cdot)$ is implemented with a sequence of convolutional (Conv), batch normalization (BN) and ReLU layers. An important detail is that the addition of~(\ref{equ:identity1}) is computed {\it between\/} the last BN and ReLU layers in the module, as shown at the top and middle of Figure~\ref{fig:bottleneck}, i.e. (\ref{equ:identity1}) is effectively implemented as
\begin{equation}
    F^l(\mathbf{x}) = ReLU\left(R(F^{l-1}(\mathbf{x})\right) + F^{l-1}(\mathbf{x})).
\label{equ:identity2}
\end{equation}

This design choice contributes to the existence of large activation maxima for deeper layers. Note that $F^{l-1}(\mathbf{x})$ is the output of a ReLU layer, i.e.
a positive number, and $R_i(F^{l-1})$, a real number. It follows that, for any $i$, there are at least two ways in which $F_i^{l}(\mathbf{x})$ can be large: 1) a large residual $R_i(F^{l-1}(\mathbf{x}))$, or 2) a positive residual $R_i(F^{l-1}(\mathbf{x}))$ if  $F_i^{l-1}(\mathbf{x})$ is already large. The second condition is particularly prone to creating large peaks, since positive residuals enable $F_i(\mathbf{x})$ to grow from layer to layer. In fact, the only way to cancel a large $F_i^{l-1}(\mathbf{x})$ is to make $R_i(F^{l-1}(\mathbf{x}))$ {\it large and negative.\/} However, it may be impossible to generate a large negative residual for one channel without generating large positive residuals for others. Hence, the attempt to correct for the second condition could create the first. This would imply that, once a large activation emerges at an intermediate layer, the network could be forced into a game of "whack-a-mole," producing large amplitudes for the subsequent layers. 

To investigate this hypothesis, we tracked the evolution of activations through the network. For a randomly selected style image, we randomly sampled image positions and tracked the corresponding activation values accross the network layers, using nearest-neighbor interpolation. Figure \ref{fig:value_evalution} shows a typical random sample of 10 activation tracks. The "whack-a-mole" effect is visible even in this limited sample. The network attempts to mitigate the peaks that appear in intermediate layers, but this creates larger picks on other channels of the subsequent layers. Note how the maximum response increases with layer depth.
Figure \ref{fig:value_evalution} also shows that, beyond the large picks, there are many small activations in the deeper layers. This is also consistent with the "whack-a-mole" hypothesis.
When combined with the ReLU of (\ref{equ:identity2}), the impetus to produce large negative residuals results in many activations $F^{l}(\mathbf{x})$ of small or zero value. Hence, as depth $l$ increases, the activations become peakier and of lower entropy.  From~(\ref{equ:gram}), the Gram matrices are then likely to have the same property. 

It should be noted that this discussion applies to the standard residual connection structure of the ResNet~\cite{he2016deep}, which is widely used. One possibility to improve stylization performance would be to change the architecture.  An example is given at the bottom of Figure \ref{fig:bottleneck}, where the skip connection is moved to precede the output ReLU of layer $l-1$. Since, in this case, the addition of activation and residual operates on two real numbers, the pressure for activations to grow would be smaller. We have tried several of these variations, but found that they tend to result in ineffective architectures for image recognition. The example modification of Figure \ref{fig:bottleneck} reduces the mean classification accuracy of the ResNet-50 on CIFAR-100 (ImageNet) from $77\%$ to $65\%$ ($76\%$ to $62\%$). Hence, even if it turned out to enhance stylization performance, it would result on different architectures for stylization and classification, which is undesirable. In what follows, we show that it is possible to improve stylization {\it while maintaining the widely used network architecture\/}.

\begin{figure*}
\centering
        \begin{subfigure}[b]{0.13\textwidth}
              \includegraphics[width=0.902\linewidth]{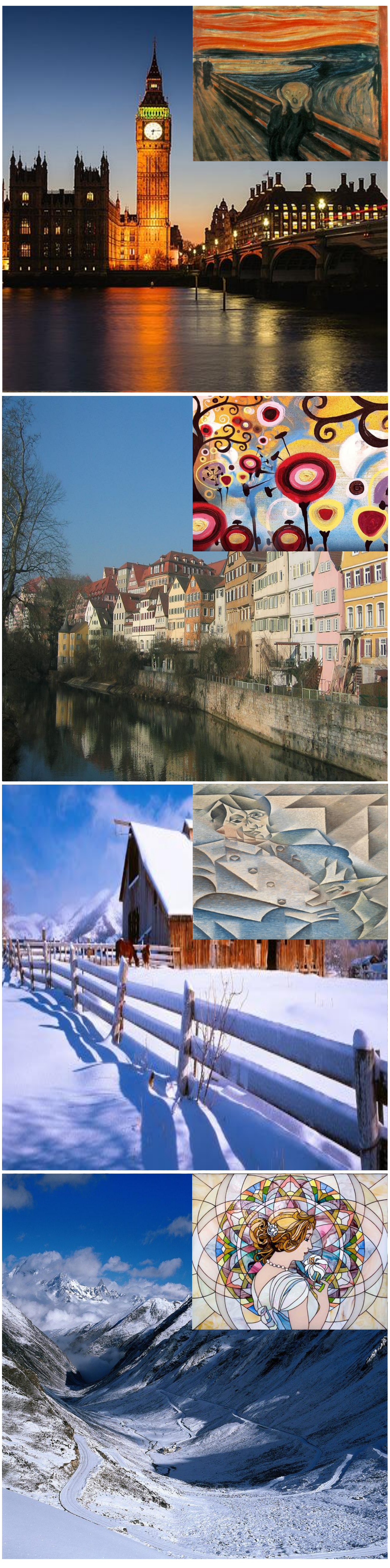}
              \caption{$\text{content}^{\text{style}}$}
                \label{fig:c_s_2}
        \end{subfigure}%
        \begin{subfigure}[b]{0.24\textwidth}
              \includegraphics[width=\linewidth]{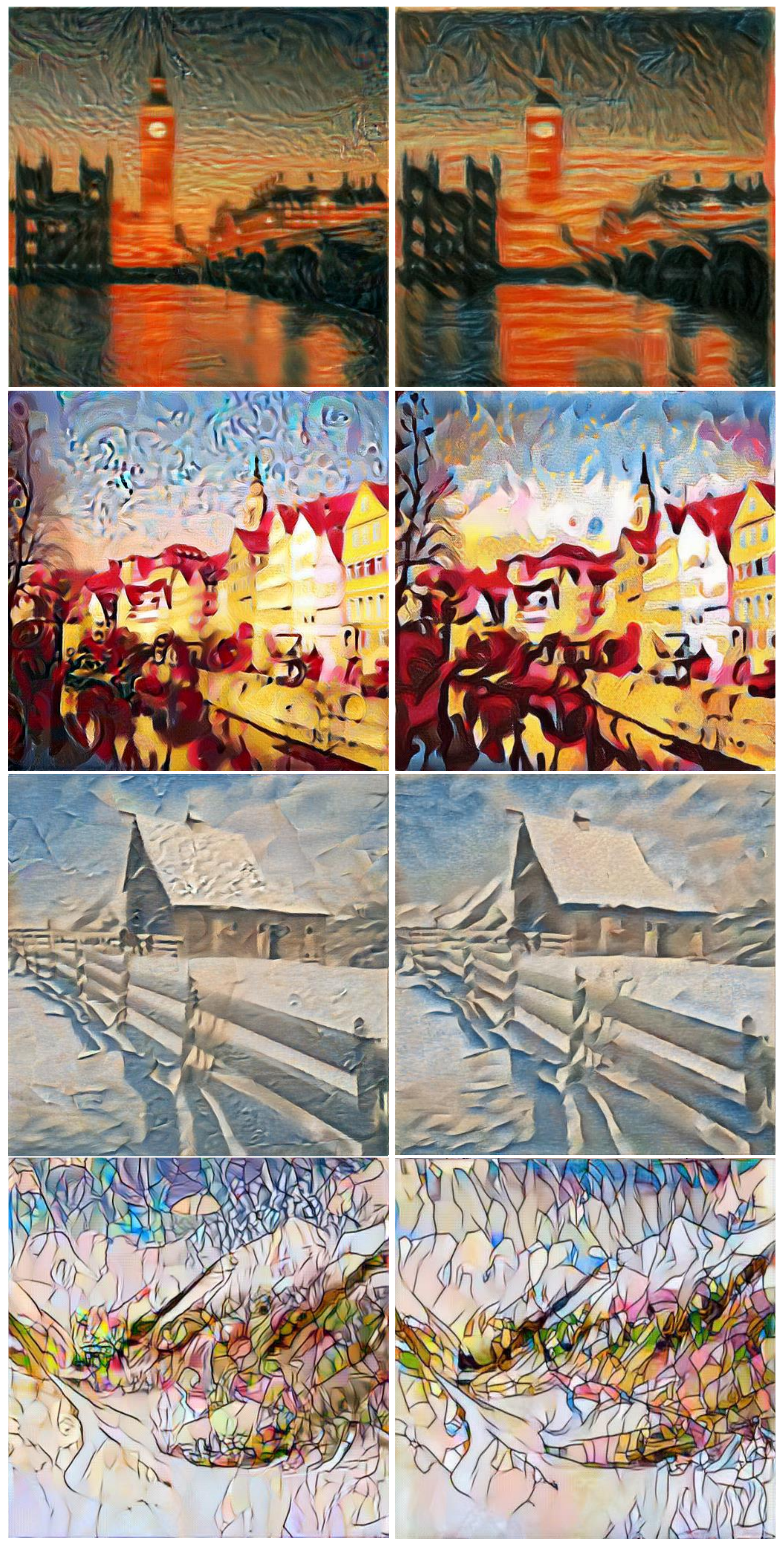}
                \caption{p-R vs p-R$^*$}
                \label{fig:pR2pRs}
        \end{subfigure}%
        \begin{subfigure}[b]{0.24\textwidth}
              \includegraphics[width=\linewidth]{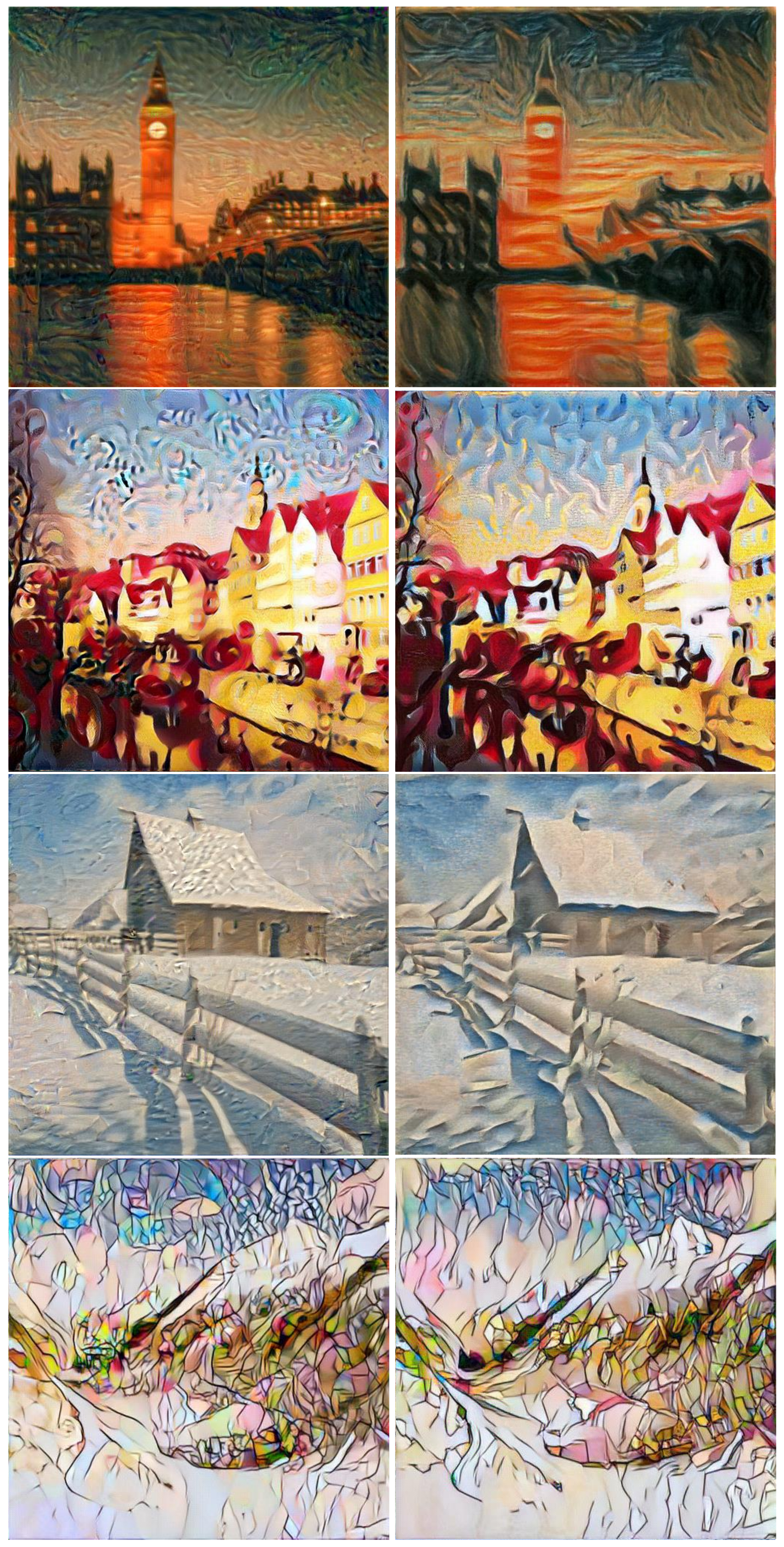}
                \caption{p-W vs p-W$^*$}
                \label{fig:pW2pWs}
        \end{subfigure}%
        \begin{subfigure}[b]{0.24\textwidth}
              \includegraphics[width=\linewidth]{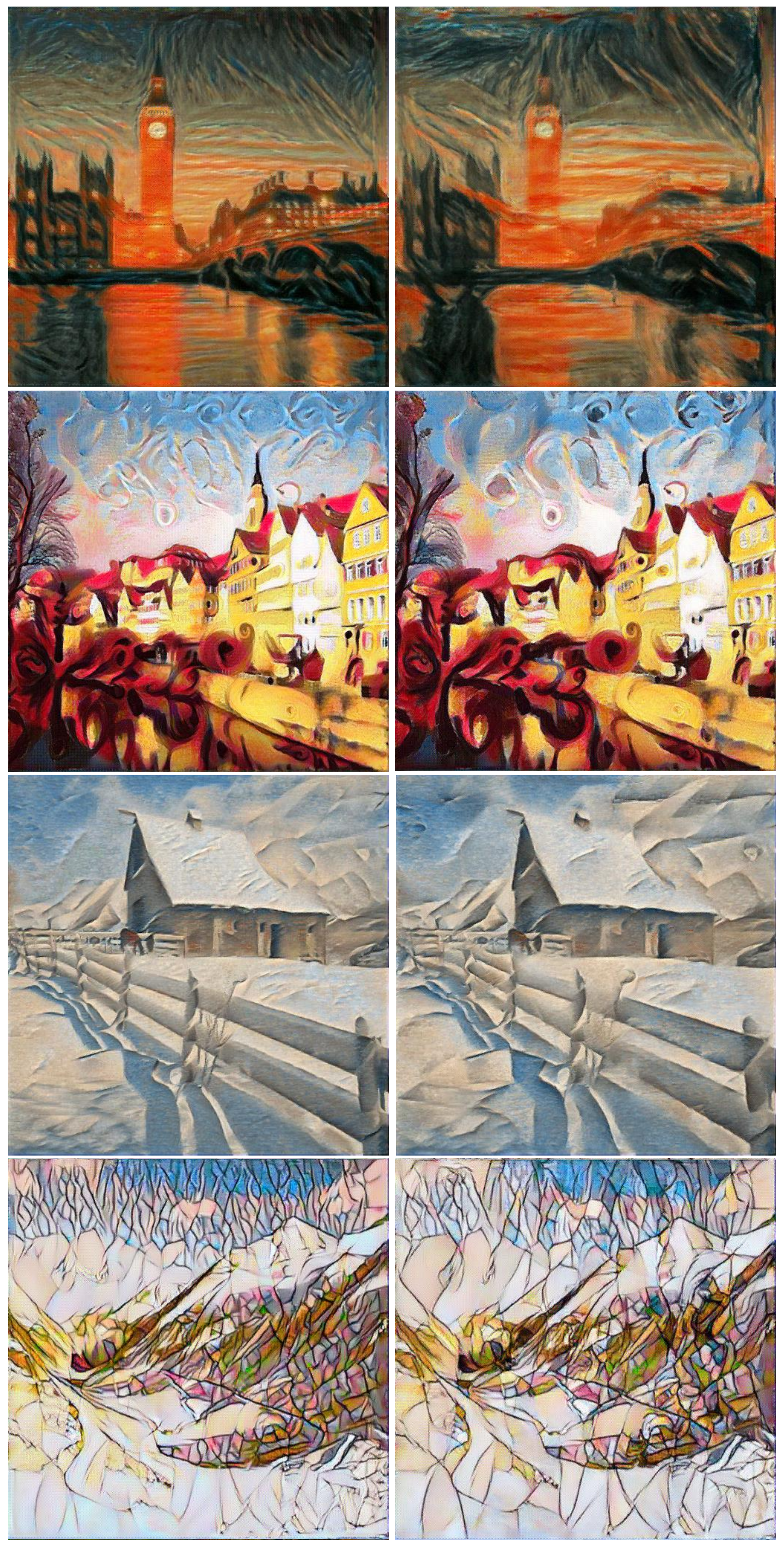}
                \caption{p-I vs p-I$^*$}
                \label{fig:pI2pIs}
        \end{subfigure}%
        \begin{subfigure}[b]{0.12\textwidth}
              \includegraphics[width=\linewidth]{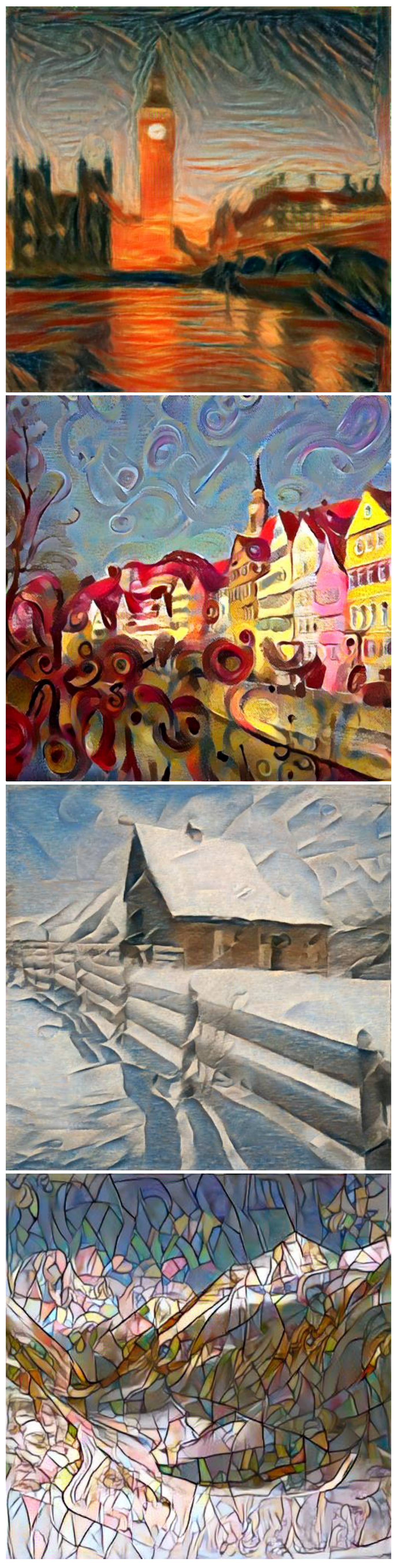}
                \caption{p-VGG}
                \label{fig:vgg}
        \end{subfigure}
        \caption{Comparison of neural style transfer performance between standard and SWAG  (denoted with $^*$) models, for different pre-trained architectures with shortcut connections (R: ResNet, W: WRN, I: Inception). The results of the standard VGG model are also shown for comparison.}\label{fig:style2}
\end{figure*}

\subsection{Stylization With Activation Smoothing}

\begin{table*}[h]
\setlength{\abovecaptionskip}{-2.0pt}
\setlength{\tabcolsep}{2pt}
\small
\begin{center}
\begin{tabular}{|l|c|c|c|c|c|c|c|}
\hline
Arch. & p-R / p-R$^*$ & p-I / p-I$^*$ &p-W / p-W$^*$& p-V / p-V$^*$&p-V / p-R$^*$& p-V / p-I$^*$&p-V / p-W$^*$\\
\hline
Pref.(\%) & 17.0 / {\bf 83.0}$_{7.1, 4.3}$ & 34.7 / {\bf65.3}$_{6.2, 5.4}$ & 32.3 / {\bf67.7}$_{3.4, 5.3}$&48.3 / {\bf51.7}$_{1.2, 5.5}$ & 33.3 / {\bf66.7}$_{3.7, 5.3}$ & 34.7 / {\bf65.3}$_{4.2, 5.4}$& 37.3 / {\bf62.7}$_{3.1, 5.5}$\\
\hline
Arch. & r-R / r-R$^*$ & r-I / r-I$^*$ &r-W / r-W$^*$&r-V / r-V$^*$ &r-V / r-R$^*$&r-V / r-I$^*$&r-V / r-W$^*$\\
\hline
Pref.(\%) & 15.7 / {\bf84.3}$_{4.6, 4.1}$ & 10.7 / {\bf89.3}$_{2.3, 3.5}$ & 5.4 / {\bf94.6}$_{5.3, 2.6}$& 44.3 / {\bf55.7}$_{2.3, 6.2}$ & 23.3 / {\bf76.7}$_{6.7, 4.8}$ &18.7 / {\bf81.3}$_{2.1, 4.4}$ & 25.3 / {\bf74.7}$_{3.4, 4.9}$\\
\hline
\end{tabular}
\end{center}
\caption{Comparison of user preference ($\%$), mean$_{\text{std, confidence interval}}$ (conf. interval at 95$\%$ conf. level). (R: ResNet; I: Inception-v3; W: WRN; V: VGG. r-: random; p-: pre-trained. *: SWAG).}
\label{tab:user_study}
\end{table*}

In this work, we leverage the observations above to improve the stylization performance of networks with residual connections. We propose a very simple solution,
inspired by the interpretation of stylization as knowledge distillation~\cite{hinton2015distilling}, where significant gains are observed by smoothing teaching distributions. In the same vein, we propose to avoid peaky activations of low entropy, by smoothing all activations with a softmax-based smoothing transformation~\footnote{We experimented with adding a temperature parameter, but this made no difference. The detailed discussion can be found in the supp.}
\begin{equation}
    \sigma(F_{ik}^l(\mathbf{x})) = \frac{e^{F_{ik}^l(\mathbf{x})}}{\sum_{m,n} e^{F_{mn}^l(\mathbf{x})}},
    \label{eq:softmax}
\end{equation}
Note that the softmax layer is {\it not\/} inserted in the network, which continues to be the original model, but only used to redefine the style and content loss functions of~(\ref{equ:stylization_loss}), which become
\begin{align}
  \mathcal{L}_{\text{content}}(\mathbf{x}^c_0, \mathbf{x})
    &= \frac{1}{2}||\sigma(F^l(\mathbf{x}))-\sigma(F^l(\mathbf{x}_0^c)) ||^2_2 \label{equ:content_loss2},\\
   \begin{split}
    \mathcal{L}_{\text{style}}(\mathbf{x}^s_0, \mathbf{x})
    &= \sum^L_l \frac{w_l}{4D^2_lM^2_l} ||G^l(\sigma(F^l(\mathbf{x})))\\
    &\quad-G^l(\sigma(F^l(\mathbf{x}_0^s))||_2^2. \label{equ:style_loss2}
   \end{split}
\end{align}
The softmax transformation reduces large peaks and increases small values, creating a more uniform distribution. Since this can be seen as a form of smoothing, we denote this approach to stylization as {\it Stylization With Activation smoothinG} (SWAG), and denote the resulting models with a `$*$' superscript, e.g. r-ResNet$^*$. ResNet$^*$ of figure \ref{fig:feature} shows that SWAG successfully suppresses the maxima and increases entropies, especially on deeper layers.

The impact of activation smoothing on stylization performance is illustrated in Figure \ref{fig:r_resnet_s} and \ref{fig:p_resnet_s}, where SWAG results are presented for the r-ResNet$^*$ and  p-ResNet$^*$ networks. In both cases, the quality of the stylized images improves substantially after smoothing,
in the sense that more high-level style patterns are transferred. The results of the r-ResNet$^*$ approach those of the r-VGG and it could be claimed that the p-ResNet$^*$ outperforms the p-VGG. This is next evaluated quantitatively.
It should be noted that there are other ways of smoothing activations and decreasing their entropy, e.g. softmax with different temperature, nested softmax, or even multiplying a small constant ($<0.1$), that we found working in our experiments. We chose the softmax of (\ref{eq:softmax}) because it is simple, hyperparameter-free, and achieves similar effects with other smoothing methods.

\vspace{0.5em}
\section{Experimental Evaluation}
\vspace{0.5em}
In this section we discuss an experimental evaluation of the stylization gains of SWAG models.
More results and implementation details can be found in the supplementary materials. 


\begin{figure*}[t]
\centering
\includegraphics[width=1.0\textwidth]{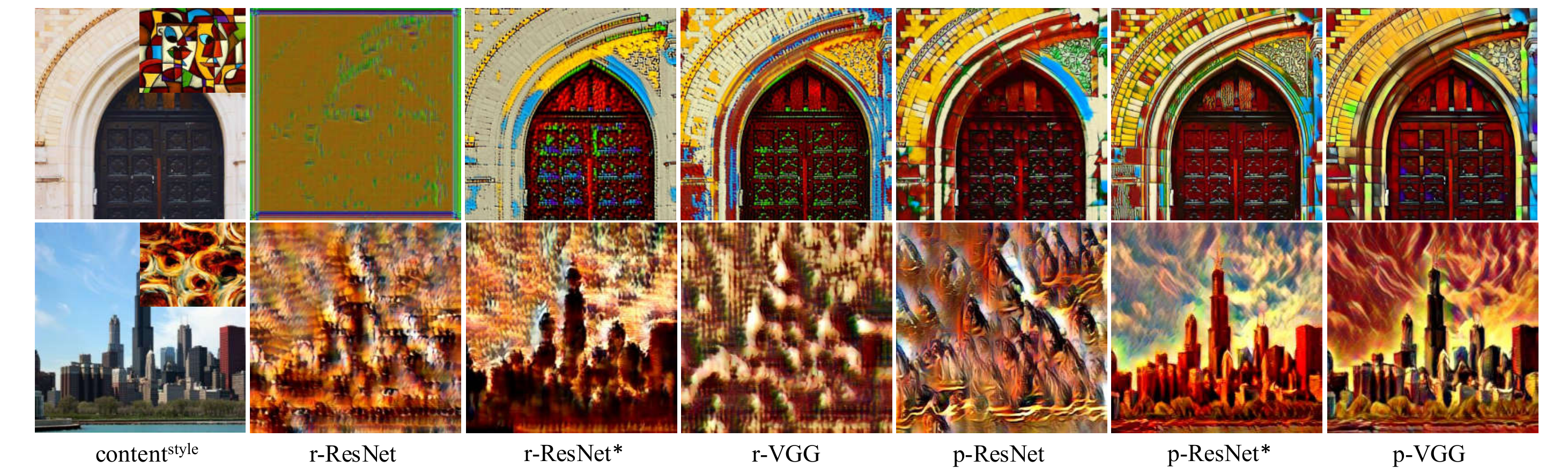}
\caption{Comparison of neural style transfer performance between standard and SWAG  (denoted with $^*$) implementations of different stylization algorithms.  Top: algorithm of~\cite{johnson2016perceptual}; Bottom: algorithm of~\cite{li2019learning}. The results of the VGG model are also shown for comparison.}
\label{fig:style3}
\end{figure*}

\vspace{0.5em}
\subsection{Qualitative evaluation} 
\vspace{0.5em}
We start by evaluating SWAG for two popular non-VGG architectures, other than the ResNet, that also rely on shortcut connections: Inception-v3~\cite{szegedy2016rethinking} and Wide ResNet (WRN)~\cite{zagoruyko2016wide}. WRN experiments use all settings of the ResNet, while conv2d$\_$1a, conv2d$\_$3b, mixed$\_$5b, mixed$\_$6a, mixed$\_$7a leayers are used for Inception. This is, again, for consistency with the VGG model of~\cite{gatys2016image}. 
We denote VGG, ResNet, WRN, and Inception-v3 networks as `\textbf{V}', `\textbf{R}', `\textbf{W}', and `\textbf{I'}, for brevity. Figure \ref{fig:style2} presents style transfer results on four different images, using the pre-trained versions of all networks, comparing results of SWAG ($^*$) and standard stylization. Note that p-R$^*$ and p-W$^*$ transfer more high-level style features, such as strokes, and textures.
For p-I$^*$, the improvement is smaller. We speculate that this is due to the somewhat different structure of the Inception whose basic module has multiple parallel connections and merges features of different solutions. This may make the stylization optimization harder. We have not investigated the issue in detail. However, the improvement is still visible. 

We next evaluate SWAG with other two stylization algorithms~\cite{johnson2016perceptual, li2019learning}. Unlike~\cite{gatys2016image}, \cite{johnson2016perceptual} trains a style-specific transformer to directly transfer the target to the stylized image, using a perceptual loss. \cite{li2019learning}, which trains a  single transformer for all types of styles, is the state-of-the-art for universal stylization. Figure \ref{fig:style3} compares the results of SWAG implementations of the two algorithms, for different networks. For both random and pre-trained models, the performance of the two algorithms improves significantly under SWAG. For example, \cite{li2019learning} fails for the p-ResNet, producing many repeated image patches. SWAG improves its results to a level comparable with VGG. These results show that SWAG is generally beneficial for stylization algorithms.


\vspace{0.5em}
\subsection{Quantitative evaluation} 
\vspace{0.5em}
Stylization quality is difficult to evaluate quantitatively, since it is subjective. While some works only show synthesized images~\cite{gatys2016image,du2020much,he2016deep,wang2017multimodal,chen2017stylebank,gatys2017controlling}, the user study has been mainly used for quantitative evaluation in this paper, where humans choose a preferred image among a set of candidates. This is consistent with the subjective nature of stylization.

We present results of a user study on Amazon Mechanical Turk, comparing pairs of images synthesized by networks with and without SWAG. Each comparison was assigned to $30$ turkers. Each turker was asked to choose the image that more closely resembled a given style image.  $10$ randomly selected stylization comparisons were performed by turker. The results of the experiment are summarized in Table \ref{tab:user_study}, which shows several interesting findings. First, for both pre-trained and random networks, models with SWAG always earned more preferences than models without. Second, all models with SWAG significantly outperformed the standard VGG implementation. All these results suggest that SWAG eliminated the dependence of stylization on the VGG architecture.

\vspace{0.5em}
\subsection{Ablation study}
\vspace{0.5em}
For stylization, the target image is usually initialized with the content image~\cite{gatys2016image,kun2016a,du2020much}, i.e. the optimization of  (\ref{equ:stylization_loss}) uses the content image $x^c_0$ as initial condition. This reduces the difficulty in transferring content information to the target image. In fact, it can make the content loss negligible~\cite{reiichiro}, to the point where the latter can be removed from the optimization altogether \cite{du2020much}. 
This creates difficulties to ablate the effectiveness of SWAG on the content and style loss individually.

\begin{figure}[t]
\centering
\includegraphics[width=0.48\textwidth]{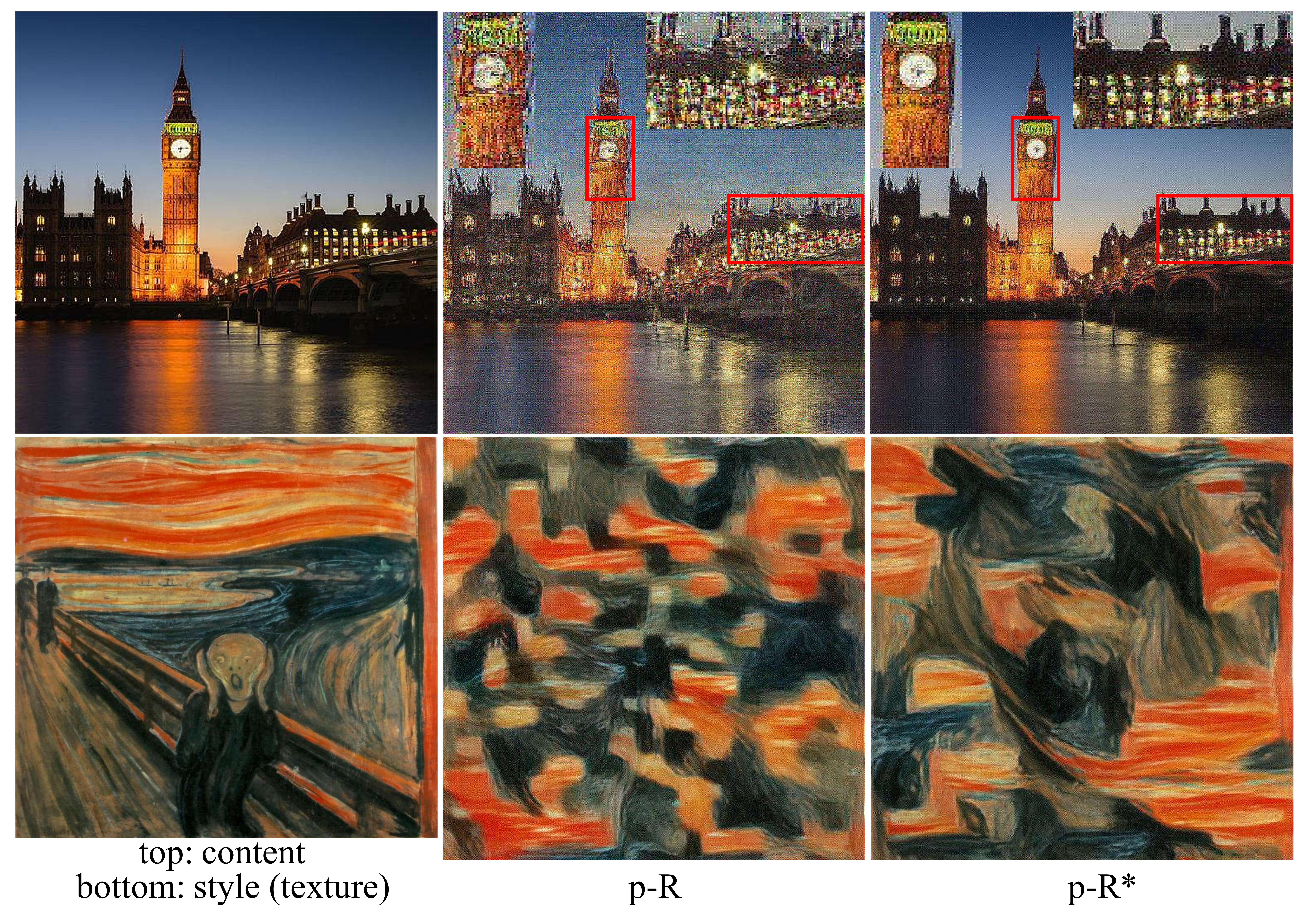}
\caption{Top: reconstruction of the top-left content image; Bottom: texture synthesis of the bottom-left image.}
\label{fig:reconst_synthesis}
\end{figure}

For this, we leveraged two alternative tasks: image reconstruction and texture synthesis, using a random initial image. Following \cite{gatys2016image,gatys2015texture}, we matched conv3$\_$4 features from the content images for image reconstruction, and multilayer features from the texture image for texture synthesis. As shown in Figure \ref{fig:reconst_synthesis}, SWAG produces better reconstructed images (average PSNR over $10$ randomly selected content images of p-R/p-R$^*$: $27.9(\pm 0.1)/28.3(\pm 0.1))$ and synthesizes textures with a larger diversity of patterns, including small-scale and large-scale style patch patterns. This suggests that 1) residual connections hurt both losses fail; 2) SWAG can match styles at deeper layers, which capture larger-scale features, to improve reconstruction and texture synthesis quality.

\section{Conclusion}

We have studied the lack of robustness of stylization algorithms for non-VGG architectures. This showed that a significant factor is the use of residual connections, which decreases the entropy of deeper layer activations. A simple solution was proposed by adding activation smoothing to the loss functions used for stylization, using a softmax function.
This was denoted as SWAG, and shown to be effective for various architectures, forms of pre-training (random vs. ImageNet), and stylization algorithms. It was shown that, with the addition of SWAG, the lightweight non-VGG becomes a viable alternative to VGG in future stylization work.

\section*{Acknowledgement} This work was partially funded by NSF awards IIS-1924937, IIS-2041009, a gift from Amazon, a gift from Qualcomm, and NVIDIA GPU donations. We also acknowledge and thank the use of the Nautilus platform for some of the experiments discussed above.

{\small
\bibliographystyle{ieee_fullname}
\bibliography{egbib}
}

\end{document}


\title{Rethinking and Improving the Robustness of Image Style Transfer \\ Supplementary Material}

\author{Pei Wang\\
UC, San Diego\\
{\tt\small pew062@ucsd.edu}
\and
Yijun Li\\
Adobe Research\\
{\tt\small yijli@adobe.com}

\and
Nuno Vasconcelos\\
UC, San Diego\\
{\tt\small nuno@ucsd.edu}
}

\maketitle

\section{Ablation study}

To validate that the poor performance of the ResNet is mainly due to the residual connection, we perform the ablation study over other network components and their combinations: (1) \textbf{VGG-bn}: VGG with batch normalization (bn) where the \emph{bn} layer is not used in the original VGG but is introduced in ResNet.
(2) \textbf{VGG-c7}: Replacing the first layer conv3$\times$3 in VGG with that conv7$\times$7 in ResNet. 
There are three kinds of conv1$\times$1 kernels in ResNet,
conv1$\times$1 that increases the channel number, conv1$\times$1 that decreases the channel number, and conv1$\times$1 that maintains the channel number, i.e. conv2$\_$1.
We denote them by `Ic1', `Dc1', `c1' respectively.
This introduces (3) \textbf{VGG-Ic1}: VGG with conv1x1 that increases the channel number, (4) \textbf{VGG-Dc1}: VGG with conv1x1 that decreases the channel number, and (5) \textbf{VGG-c1}: VGG with conv1x1 that maintains the channel number. 
We also investigate the influence of the combination of many factors. (6) \textbf{VGG-Ic1-Dc1}, (7) \textbf{VGG-c7-Ic1}, (8) \textbf{VGG-c7-Dc1}, (9) \textbf{VGG-c7-Ic1-Dc1}. The impact of other factors like the number of channels per layer or network depth have been discussed and shown to be 
less important in~\cite{du2020much}.
Figure \ref{fig:ablation} presents two stylization examples obtained by models (1)$\sim$(9). It can be seen that the influence brought by those network components are much smaller than that by the residual connection (as shown in Figure 2 of the paper).
Therefore, we conclude that the residual connection in ResNet is the root cause which results in the poor stylization performance.

\section{More results}

\subsection{Style loss}

\begin{table*}[th!]
\setlength{\abovecaptionskip}{-2.0pt}
\setlength{\tabcolsep}{2pt}
\begin{center}
\begin{tabular}{|l|c|c|c|c||c|c|c|c|}
\hline
& \multicolumn{4}{|c||}{Pre-trained} & \multicolumn{4}{|c|}{Random}\\
\hline
Arch. & ResNet & Inception & WRN&VGG& ResNet & Inception & WRN & VGG\\
\hline
Standard & 3.8(9.1)e4 & 6.3(3.4)e4 & 3.8(4.5)e4 & 2.5(4.6)e4 & 1.8(1.3)e6 & 1.3(9.7)e6 &1.3(7.7)e6 &3.9(7.1)e4\\
SWAG & {\bf 2.3(4.3)e4} & {\bf4.0(1.9)e4} & {\bf2.6(6.3)e4} & {\bf2.4(4.6)e4} &{\bf 3.4(1.3)e4} & {\bf7.9(6.6)e4}&{\bf6.3(3.6)e4} & {\bf3.7(7.0)e4}\\
\hline
\end{tabular}
\end{center}
\caption{Style loss comparison of different architectures (mean(std)).}
\label{tab:style_loss}
\end{table*}

We also compute the style loss for images synthesized by a pre-specified model (usually a pre-trained VGG)~\cite{li2017universal,johnson2016perceptual}.
Specifically, images are stylized using the different network architectures, with and without SWAG. Stylized images are then fed into a VGG pre-trained on ImageNet and the loss of (4) is computed.
This measures the similarity between synthesized and style image, ignoring content information. Note that this metric has a certain bias towards the VGG.

Table \ref{tab:style_loss} shows the results of the style loss comparison, based on activations from five layers (conv1$\_$1, conv2$\_$1, conv3$\_$1, conv4$\_$1, conv5$\_$1) of the pre-trained VGG\footnote{The models, pre-trained on ImageNet, are those provided by PyTorch  (\url{https://pytorch.org/docs/stable/torchvision/models.html}).}. 
We randomly select $10$ content images and $10$ style images from~\cite{gatys2016image,du2020much,li2018closed}, and compute the averaged style loss over all $100$ content-style combinations. A few conclusions can be drawn. First, for both pre-trained and random models, SWAG improves the performance of each non-VGG network. Second, for random networks, the gains of SWAG are of two orders of magnitude. Third, for both random and pre-trained models, the ResNet with SWAG even outperforms the standard VGG model. 
Fourth, SWAG even slightly improves the performances of the VGG model, which does not suffer from a noticeable peaky large activation problem. Finally, SWAG significantly reduces the performance gap between random and pre-trained models.

\subsection{Visual comparison}

More comparisons of images synthesized by different networks corresponding to the $12$ comparison pairs in Table 2 of the paper are shown in Figure 3$\sim$14.

\section{Ablation study on $T$ in Eq. (11) of the paper}


In general, $T$ should 1) increase as $H$ in Eq. (5) decreases and 2) be $\geq 1$ (note that $H \in [0, 1]$). This is to guarantee SWAG can always increase the entropy and be more powerful for ultra-peaky activations.
The ablation study experiment on temperature ($T$) in Eq. (11) is conducted. 
We found that mean entropy increases with $T$ but saturates quickly for $T > 1$. The mean style loss across different architectures also decreases until saturation for $T > 1$. It indicates  
$T=1$ is sufficient and larger $T$ will not result in further improvement. 

        

\section{Implementation details}

The content and style image are subject to the standard normalization. 
Specifically, all images are first converted to $[0.0, 1.0]$ from $[0, 255]$ and then normalized by subtracting the mean $([0.485, 0.456, 0.406])$ and divided by the standard deviation $(0.229, 0.224, 0.225])$ of each RGB color channel. All results are of size $512 \times 512$. All pre-trained models used in the paper are from PyTorch\footnote{ \url{https://pytorch.org/docs/stable/torchvision/models.html}}. 
We follow the setup of \cite{gatys2016image} for the VGG model, i.e., using features at the conv1$\_$1, conv2$\_$1, conv3$\_$1, conv4$\_$1, conv5$\_$1 layer for style loss of Eq. (4) and Eq. (13), and the conv4$\_$2 layer for content loss of Eq. (3) and Eq. (12). 
We set $\alpha=1$ and $\beta=4e10$ in Eq. (2). For ResNet, we follow the setting of \cite{reiichiro}, but, in addition to the features at the conv2$\_$3, conv3$\_$4, conv4$\_$6, conv5$\_$3 layer, we also use the conv1$\_$2 layer in Eq. (4) and Eq. (13). This is for fair comparison with the VGG implementation of~\cite{gatys2016image}, which uses five layers at different scales. 
The conv4$\_$6 layer is used for the computation of content loss. We set $\alpha=1$ and $\beta=1e17$. The same setting is for WRN. 
On Inception v3, the conv2d$\_$1a, conv2d$\_$3b, mixed$\_$5b, mixed$\_$6a, mixed$\_$7a leayers are used, again for consistency with the VGG model. The mixed$\_$5b layer is for content loss computation, and we set $\alpha=1$, $\beta=4e10$.

\begin{figure*}
        \begin{subfigure}[b]{0.199\textwidth}
             \includegraphics[width=\linewidth]{figures_supp/content_style1.pdf}
              \caption{$\text{content}^{\text{style}}$}
                \label{fig:c_s_1}
        \end{subfigure}%
        \begin{subfigure}[b]{0.2\textwidth}
             \includegraphics[width=\linewidth]{figures_supp/r_vgg_bn.pdf}
                \caption{r-VGG-bn}
                \label{fig:r-VGG-bn}
        \end{subfigure}%
        \begin{subfigure}[b]{0.2\textwidth}
             \includegraphics[width=\linewidth]{figures_supp/r_vgg_conv1_less.pdf}
                \caption{r-VGG-Ic1}
                \label{fig:r-VGG-c1_2}
        \end{subfigure}%
        \begin{subfigure}[b]{0.2\textwidth}
              \includegraphics[width=\linewidth]{figures_supp/r_vgg_conv1.pdf}
                \caption{r-VGG-Dc1}
                \label{fig:r-VGG-c1}
        \end{subfigure}%
        \begin{subfigure}[b]{0.2\textwidth}
             \includegraphics[width=\linewidth]{figures_supp/r_vgg_iconv1.pdf}
                \caption{r-VGG-c1}
                \label{fig:r-VGG-Ic1}
        \end{subfigure}%
        
        \begin{subfigure}[b]{0.199\textwidth}
             \includegraphics[width=\linewidth]{figures_supp/r_vgg_conv7.pdf}
              \caption{r-VGG-c7}
                \label{fig:r-VGG-c7}
        \end{subfigure}%
        \begin{subfigure}[b]{0.2\textwidth}
             \includegraphics[width=\linewidth]{figures_supp/r_vgg_conv1_iconv1.pdf}
                \caption{r-VGG-Ic1-Dc1}
                \label{fig:r-VGG-c1-Ic1}
        \end{subfigure}%
        \begin{subfigure}[b]{0.2\textwidth}
             \includegraphics[width=\linewidth]{figures_supp/r_vgg_conv7_conv1.pdf}
                \caption{r-VGG-c7-Ic1}
                \label{fig:r-VGG-c7-c1}
        \end{subfigure}%
        \begin{subfigure}[b]{0.2\textwidth}
             \includegraphics[width=\linewidth]{figures_supp/r_vgg_conv7_iconv1.pdf}
                \caption{r-VGG-c7-Dc1}
                \label{fig:r-VGG-c7-Ic1}
        \end{subfigure}%
        \begin{subfigure}[b]{0.2\textwidth}
             \includegraphics[width=\linewidth]{figures_supp/r_vgg_conv7_conv1_iconv1.pdf}
                \caption{r-VGG-c7-Ic1-Dc1}
                \label{fig:r-VGG-c7-c1-Ic1}
        \end{subfigure}%
        \caption{Results comparisons of different architectures. (`r-' represent randomly initialized. Detailed comparison need to zoom in on pictures. `$^*$' denotes our proposal.)}\label{fig:ablation}
\end{figure*}

\begin{figure*}
\centering
        \begin{subfigure}[b]{0.3\textwidth}
              \includegraphics[width=\linewidth]{figures_supp/pR2pRs_c_s.pdf}
              \caption{$\text{content}^{\text{style}}$}
                \label{fig:c_s}
        \end{subfigure}%
        \begin{subfigure}[b]{0.3\textwidth}
              \includegraphics[width=\linewidth]{figures_supp/pR2pRs_pR.pdf}
                \caption{p-R}
                \label{fig:rR}
        \end{subfigure}%
        \begin{subfigure}[b]{0.3\textwidth}
              \includegraphics[width=\linewidth]{figures_supp/pR2pRs_pRs.pdf}
                \caption{p-R$^*$}
                \label{fig:rRs}
        \end{subfigure}
        \caption{Comparison of neural style transfer performance between p-R and p-R SWAG  (denoted with $^*$) models}\label{fig:style2}
\end{figure*}

\begin{figure*}
\centering
        \begin{subfigure}[b]{0.3\textwidth}
              \includegraphics[width=\linewidth]{figures_supp/pI2pIs_c_s.pdf}
              \caption{$\text{content}^{\text{style}}$}
                \label{fig:c_s}
        \end{subfigure}%
        \begin{subfigure}[b]{0.3\textwidth}
              \includegraphics[width=\linewidth]{figures_supp/pI2pIs_pI.pdf}
                \caption{p-I}
                \label{fig:rR}
        \end{subfigure}%
        \begin{subfigure}[b]{0.3\textwidth}
              \includegraphics[width=\linewidth]{figures_supp/pI2pIs_pIs.pdf}
                \caption{p-I$^*$}
                \label{fig:rRs}
        \end{subfigure}
        \caption{Comparison of neural style transfer performance between p-I and p-I SWAG  (denoted with $^*$) models}\label{fig:style2}
\end{figure*}

\begin{figure*}
\centering
        \begin{subfigure}[b]{0.3\textwidth}
              \includegraphics[width=\linewidth]{figures_supp/pW2pWs_c_s.pdf}
              \caption{$\text{content}^{\text{style}}$}
                \label{fig:c_s}
        \end{subfigure}%
        \begin{subfigure}[b]{0.3\textwidth}
              \includegraphics[width=\linewidth]{figures_supp/pW2pWs_pW.pdf}
                \caption{p-W}
                \label{fig:rR}
        \end{subfigure}%
        \begin{subfigure}[b]{0.3\textwidth}
              \includegraphics[width=\linewidth]{figures_supp/pW2pWs_pWs.pdf}
                \caption{p-W$^*$}
                \label{fig:rRs}
        \end{subfigure}
        \caption{Comparison of neural style transfer performance between p-W and p-W SWAG  (denoted with $^*$) models}\label{fig:style2}
\end{figure*}

\begin{figure*}
\centering
        \begin{subfigure}[b]{0.3\textwidth}
              \includegraphics[width=\linewidth]{figures_supp/rR2rRs_c_s.pdf}
              \caption{$\text{content}^{\text{style}}$}
                \label{fig:c_s}
        \end{subfigure}%
        \begin{subfigure}[b]{0.3\textwidth}
              \includegraphics[width=\linewidth]{figures_supp/rR2rRs_rR.pdf}
                \caption{r-R}
                \label{fig:rR}
        \end{subfigure}%
        \begin{subfigure}[b]{0.3\textwidth}
              \includegraphics[width=\linewidth]{figures_supp/rR2rRs_rRs.pdf}
                \caption{r-R$^*$}
                \label{fig:rRs}
        \end{subfigure}
        \caption{Comparison of neural style transfer performance between r-R and r-R SWAG  (denoted with $^*$) models}\label{fig:style2}
\end{figure*}

\begin{figure*}
\centering
        \begin{subfigure}[b]{0.3\textwidth}
              \includegraphics[width=\linewidth]{figures_supp/rI2rIs_c_s.pdf}
              \caption{$\text{content}^{\text{style}}$}
                \label{fig:c_s}
        \end{subfigure}%
        \begin{subfigure}[b]{0.3\textwidth}
              \includegraphics[width=\linewidth]{figures_supp/rI2rIs_rI.pdf}
                \caption{r-I}
                \label{fig:rR}
        \end{subfigure}%
        \begin{subfigure}[b]{0.3\textwidth}
              \includegraphics[width=\linewidth]{figures_supp/rI2rIs_rIs.pdf}
                \caption{r-I$^*$}
                \label{fig:rRs}
        \end{subfigure}
        \caption{Comparison of neural style transfer performance between r-I and r-I SWAG  (denoted with $^*$) models}\label{fig:style2}
\end{figure*}

\begin{figure*}
\centering
        \begin{subfigure}[b]{0.3\textwidth}
              \includegraphics[width=\linewidth]{figures_supp/rW2rWs_c_s.pdf}
              \caption{$\text{content}^{\text{style}}$}
                \label{fig:c_s}
        \end{subfigure}%
        \begin{subfigure}[b]{0.3\textwidth}
              \includegraphics[width=\linewidth]{figures_supp/rW2rWs_rW.pdf}
                \caption{r-W}
                \label{fig:rR}
        \end{subfigure}%
        \begin{subfigure}[b]{0.3\textwidth}
              \includegraphics[width=\linewidth]{figures_supp/rW2rWs_rWs.pdf}
                \caption{r-W$^*$}
                \label{fig:rRs}
        \end{subfigure}
        \caption{Comparison of neural style transfer performance between r-W and r-W SWAG  (denoted with $^*$) models}\label{fig:style2}
\end{figure*}


\begin{figure*}
\centering
        \begin{subfigure}[b]{0.3\textwidth}
              \includegraphics[width=\linewidth]{figures_supp/pV2pRs_c_s.pdf}
              \caption{$\text{content}^{\text{style}}$}
                \label{fig:c_s}
        \end{subfigure}%
        \begin{subfigure}[b]{0.3\textwidth}
              \includegraphics[width=\linewidth]{figures_supp/pV2pRs_pV.pdf}
                \caption{p-V}
                \label{fig:rR}
        \end{subfigure}%
        \begin{subfigure}[b]{0.3\textwidth}
              \includegraphics[width=\linewidth]{figures_supp/pV2pRs_pRs.pdf}
                \caption{p-R$^*$}
                \label{fig:rRs}
        \end{subfigure}
        \caption{Comparison of neural style transfer performance between p-V and p-R SWAG  (denoted with $^*$) models}\label{fig:style2}
\end{figure*}

\begin{figure*}
\centering
        \begin{subfigure}[b]{0.3\textwidth}
              \includegraphics[width=\linewidth]{figures_supp/pV2pIs_c_s.pdf}
              \caption{$\text{content}^{\text{style}}$}
                \label{fig:c_s}
        \end{subfigure}%
        \begin{subfigure}[b]{0.3\textwidth}
              \includegraphics[width=\linewidth]{figures_supp/pV2pIs_pV.pdf}
                \caption{p-V}
                \label{fig:rR}
        \end{subfigure}%
        \begin{subfigure}[b]{0.3\textwidth}
              \includegraphics[width=\linewidth]{figures_supp/pV2pIs_pIs.pdf}
                \caption{p-I$^*$}
                \label{fig:rRs}
        \end{subfigure}
        \caption{Comparison of neural style transfer performance between p-V and p-I SWAG  (denoted with $^*$) models}\label{fig:style2}
\end{figure*}

\begin{figure*}
\centering
        \begin{subfigure}[b]{0.3\textwidth}
              \includegraphics[width=\linewidth]{figures_supp/pV2pWs_c_s.pdf}
              \caption{$\text{content}^{\text{style}}$}
                \label{fig:c_s}
        \end{subfigure}%
        \begin{subfigure}[b]{0.3\textwidth}
              \includegraphics[width=\linewidth]{figures_supp/pV2pWs_pV.pdf}
                \caption{p-V}
                \label{fig:rR}
        \end{subfigure}%
        \begin{subfigure}[b]{0.3\textwidth}
              \includegraphics[width=\linewidth]{figures_supp/pV2pWs_pWs.pdf}
                \caption{p-W$^*$}
                \label{fig:rRs}
        \end{subfigure}
        \caption{Comparison of neural style transfer performance between p-V and p-W SWAG  (denoted with $^*$) models}\label{fig:style2}
\end{figure*}

\begin{figure*}
\centering
        \begin{subfigure}[b]{0.3\textwidth}
              \includegraphics[width=\linewidth]{figures_supp/rV2rRs_c_s.pdf}
              \caption{$\text{content}^{\text{style}}$}
                \label{fig:c_s}
        \end{subfigure}%
        \begin{subfigure}[b]{0.3\textwidth}
              \includegraphics[width=\linewidth]{figures_supp/rV2rRs_rV.pdf}
                \caption{r-V}
                \label{fig:rR}
        \end{subfigure}%
        \begin{subfigure}[b]{0.3\textwidth}
              \includegraphics[width=\linewidth]{figures_supp/rV2rRs_rRs.pdf}
                \caption{r-R$^*$}
                \label{fig:rRs}
        \end{subfigure}
        \caption{Comparison of neural style transfer performance between r-V and r-R SWAG  (denoted with $^*$) models}\label{fig:style2}
\end{figure*}

\begin{figure*}
\centering
        \begin{subfigure}[b]{0.3\textwidth}
              \includegraphics[width=\linewidth]{figures_supp/rV2rIs_c_s.pdf}
              \caption{$\text{content}^{\text{style}}$}
                \label{fig:c_s}
        \end{subfigure}%
        \begin{subfigure}[b]{0.3\textwidth}
              \includegraphics[width=\linewidth]{figures_supp/rV2rIs_rV.pdf}
                \caption{r-V}
                \label{fig:rR}
        \end{subfigure}%
        \begin{subfigure}[b]{0.3\textwidth}
              \includegraphics[width=\linewidth]{figures_supp/rV2rIs_rIs.pdf}
                \caption{r-I$^*$}
                \label{fig:rRs}
        \end{subfigure}
        \caption{Comparison of neural style transfer performance between r-V and r-I SWAG  (denoted with $^*$) models}\label{fig:style2}
\end{figure*}

\begin{figure*}
\centering
        \begin{subfigure}[b]{0.3\textwidth}
              \includegraphics[width=\linewidth]{figures_supp/rV2rWs_c_s.pdf}
              \caption{$\text{content}^{\text{style}}$}
                \label{fig:c_s}
        \end{subfigure}%
        \begin{subfigure}[b]{0.3\textwidth}
              \includegraphics[width=\linewidth]{figures_supp/rV2rWs_rV.pdf}
                \caption{r-V}
                \label{fig:rR}
        \end{subfigure}%
        \begin{subfigure}[b]{0.3\textwidth}
              \includegraphics[width=\linewidth]{figures_supp/rV2rWs_rWs.pdf}
                \caption{r-W$^*$}
                \label{fig:rRs}
        \end{subfigure}
        \caption{Comparison of neural style transfer performance between r-V and r-W SWAG  (denoted with $^*$) models}\label{fig:style2}

\end{figure*}

{\small
\bibliographystyle{ieee_fullname}
\bibliography{egbib}
}